\documentclass[final,3p,times,authoryear]{elsarticle}

\usepackage{amsmath,amssymb}
\usepackage{graphicx}
\usepackage{subcaption}
\usepackage{graphicx}
\usepackage{booktabs}
\usepackage{multirow}
\usepackage{siunitx}
\usepackage{hyperref}
\usepackage[normalem]{ulem}
\usepackage{enumitem}
\hypersetup{
  colorlinks=true,
  linkcolor=blue,
  citecolor=blue,
  urlcolor=blue
}
\usepackage{xcolor}
\definecolor{DarkGreen}{RGB}{34, 139, 34} 
\begin{document}

\begin{frontmatter}

\title{Beyond AlphaEarth: Toward Human-Centered Geospatial Foundation Models via POI-Guided Contrastive Learning}

\author[inst1]{Junyuan Liu}
\author[inst1,inst2]{Quan Qin}
\author[inst1,inst3]{Guangsheng Dong}
\author[inst1]{Xinglei Wang}
\author[inst1]{Jiazhuang Feng}
\author[inst1,inst4]{Zichao Zeng}  

\author[inst1]{Tao Cheng\corref{cor1}}
\cortext[cor1]{Corresponding author.}
\ead{tao.cheng@ucl.ac.uk}

\affiliation[inst1]{organization={SpaceTimeLab, Department of Civil, Environmental and Geomatic Engineering, University College London}, 
            city={London}, postcode={WC1E 6BT}, country={United Kingdom}}

\affiliation[inst2]{organization={School of Resource and Environmental Sciences, Wuhan University}, 
            city={Wuhan}, postcode={430079}, country={China}}

\affiliation[inst3]{organization={State Key Laboratory of Information Engineering in Surveying, Mapping and Remote Sensing, Wuhan University}, 
            city={Wuhan}, postcode={430079}, country={China}}

\affiliation[inst4]{organization={3DIMPact, Department of Civil, Environmental and Geomatic Engineering, University College London}, 
            city={London}, postcode={WC1E 6BT}, country={United Kingdom}}  

\begin{abstract}
Recent geospatial foundation models (GFMs) produce spatially extensive representations of the Earth’s surface that capture rich physical and environmental patterns. Among them, the AlphaEarth Foundation (AE) represents a major step, generating 10\,m embeddings from multi-source Earth Observation (EO) data that includes diverse environmental and spectral characteristics.
However, such EO-driven representations primarily encode physical and spectral patterns rather than human activities or urban semantics, limiting their ability to capture the functional dimensions of cities and making the learned representations difficult to interpret or query using natural language.
We introduce \textbf{AETHER} (\textbf{A}lpha\textbf{E}ar\textbf{th}–POI \textbf{E}nriched \textbf{R}epresentation Learning), 
a lightweight framework that aligns AlphaEarth with human-centered urban analysis through multimodal alignment guided by Points of Interest (POIs). 
By enforcing both cross-modal AE–POI alignment and intra-modal multi-scale consistency, AETHER integrates functional urban semantics with EO-driven representations and grounds the embedding space in natural language.
The resulting representations support both urban mapping tasks and natural language–conditioned spatial retrieval.
Experiments across four downstream tasks in Greater London and Singapore demonstrate consistent state-of-the-art performance, with relative improvements ranging from 4.5\% to 21.9\%.
Furthermore, the aligned embedding space enables spatial localization through natural language queries. 
By aligning EO-based foundation models with human-centered semantics, AETHER improves the interpretability of geospatial representations and advances geospatial representation learning toward human-centered, language-accessible geospatial foundation models.
\end{abstract}
\begin{keyword}
Geospatial Foundation Model \sep
Urban Representation Learning \sep
Contrastive Learning \sep
Points of Interest \sep
Language Alignment
\end{keyword}

\end{frontmatter}

\section{Introduction}

Understanding the spatial organization and functional dynamics of urban systems remains a long-standing challenge in GIScience and urban computing. 
Recent studies suggest that unified spatial representations capable of generalizing across scales, modalities, and environmental contexts provide a promising direction for large-scale geographic understanding \citep{klemmer2025earth,janowicz2025geofm}.
Building on this idea, recent advances in geospatial foundation models (GFMs) have introduced a new paradigm for learning general-purpose geographic representations from large-scale Earth Observation (EO) data \citep{yan2024urbanclip, klemmer2025satclip, brown2025alphaearth}. 
Among these, the AlphaEarth Foundation model \citep{brown2025alphaearth} represents one of the most comprehensive efforts to date, producing 64-dimensional embeddings at 10\,m resolution from multi-source EO data. 
For simplicity, we refer to it as AlphaEarth (AE) hereafter. 
These embeddings excel in natural land-surface tasks such as land-cover classification and ecological mapping, demonstrating the potential of large-scale visual pretraining for spatial representation learning.

However, applying such representations to urban contexts remains challenging. 
Urban environments are shaped not only by physical form but also by human activities, institutional structures, and socioeconomic interactions \citep{batty2013new}. 
While AE provides spatially complete and globally consistent coverage, its embeddings predominantly encode spectral and environmental signals, limiting their ability to represent the functional organization of urban systems. 

Moreover, although the pretraining of some GFMs incorporates some language supervision, their embeddings remain only weakly aligned with language semantics. 
As a result, functional patterns implicitly reflected in EO features cannot be easily interpreted or queried through natural language. 
Current geospatial foundation models therefore produce dense latent representations whose semantic content remains difficult to access without task-specific supervision. 
This limited semantic accessibility reduces interpretability and makes it challenging to understand how foundation models encode human-relevant spatial functions.

Points of Interest (POIs) provide human-centered cues. 
By linking geographic locations with semantic labels and descriptive names, POIs encode both the “where” and the “what” of urban space \citep{goodchild2020platial}. 
Such textual descriptions encode rich functional information about places and human activities.
With recent advances in language models, this textual information can be embedded into representations that capture complex semantic relationships. This makes POIs a natural bridge between geographic locations and language-based functional semantics.
Prior studies have leveraged POI data to model functional regions and human activity patterns \citep{yan2017itdl, niu2021delineating, huang2022sppe}, and recent frameworks such as CaLLiPer \citep{wang2025multi, liu2025enriching} further align textual and spatial embeddings to enhance semantic expressiveness. 
However, POI-based representations suffer from uneven spatial distribution and lack explicit spatial structure.

These observations motivate the following question:
\emph{How can physically grounded EO representations be aligned with human-centered semantic signals from POIs to produce spatial representations that are both functionally expressive and language-accessible?}
To this end, we propose \textbf{AETHER} (\textbf{A}lpha\textbf{E}ar\textbf{th}--POI \textbf{E}nriched \textbf{R}epresentation Learning), a multimodal contrastive alignment framework that adapts AE for human-centered urban analysis. AETHER aggregates AE embeddings within spatial windows around POI locations, projects them through multi-scale alignment heads, and aligns them with POI text embeddings derived from pretrained language models using an InfoNCE-based objective. The aligned embeddings can then be flexibly aggregated for supervised urban mapping tasks or directly conditioned on natural language queries for spatial retrieval.

We evaluate AETHER across four urban tasks in Greater London and Singapore, covering discrete classification (land-use classification), compositional distribution modeling (socioeconomic and land-use composition mapping), and scalar regression (GDP prediction). In addition, we assess the semantic grounding capability of the learned embedding space through natural language–conditioned spatial retrieval. Across cities and tasks, AETHER consistently outperforms existing baselines while enabling intuitive language-based spatial querying.

In summary, the main contributions of this study are:
\begin{itemize}
    \item \textbf{Bridging the gap between EO-driven representations and urban semantics.}
    To the best of our knowledge, AETHER is the first framework that connects physically grounded AlphaEarth embeddings with human-centered semantic signals through POI-guided contrastive alignment.

    \item \textbf{A POI-guided multimodal alignment framework.}
    We propose a lightweight contrastive framework that aggregates AE embeddings at multiple spatial scales and aligns them with POI textual embeddings through cross-modal AE--POI alignment and intra-modal AE--AE consistency.

    \item \textbf{Enhancing interpretability of geospatial representations.}
    Through language--semantic alignment with a multi-scale spatial module, the learned embedding space becomes interpretable using natural language, enabling open-vocabulary spatial retrieval without task-specific supervision.

    \item \textbf{Comprehensive empirical validation.}  
    Experiments in London and Singapore demonstrate that AETHER achieves state-of-the-art performance across four downstream tasks, while showing strong robustness and data efficiency across diverse urban settings.
\end{itemize}

For reproducibility and further research, our implementation is publicly available at \url{https://github.com/inwind0212/AETHER}.

\section{Related Work}

\subsection{Spatial Representations from Diverse Data}

Spatial representation learning in GIScience has leveraged a variety of geospatial data sources, each capturing different aspects of geographic space and urban environments.
Depending on their origin and information content, these data give rise to distinct modeling paradigms that capture spatial configuration, physical characteristics, semantic functions, and dynamic activities from different perspectives.  
Accordingly, existing studies can be broadly grouped into three major lines: location-based, EO-based, and POI-based representations.

Location encoding methods transform geographic coordinates into continuous embeddings that can be used by neural networks \citep{mai2022review}.
Approaches such as Wrap \citep{mac2019presence}, Space2Vec \citep{mai2020iclr}, Sphere2Vec \citep{mai2023sphere2vec}, and Spherical Harmonics \citep{russwurm2024sh} generate smooth, inductive spatial representations that capture geographic continuity and spatial autocorrelation.  
While these models provide dense spatial coverage and strong generalization, they primarily encode \emph{where} places are rather than \emph{what} they mean or \emph{how} they function—leaving semantic understanding to be addressed by multimodal alignment frameworks.

EO has long served as a fundamental data source for space understanding.  
In geospatial research, large-scale pretraining frameworks, such as SatCLIP \citep{klemmer2025satclip} and GeoCLIP \citep{vivanco2024geoclip}, align satellite imagery with geographic coordinates to generate general-purpose embeddings.  
AlphaEarth \citep{brown2025alphaearth} further integrates multi-source EO data into 10\,m embeddings that capture detailed land-surface morphology and environmental context.  
However, these EO-driven models primarily encode physical and spectral features, limiting their ability to represent human functions or socioeconomic structures.

POIs provide a complementary and human-centered view of cities.  
Each POI conveys semantic information through its location, category, and name about the role of a place, such as education, commerce, or healthcare.  
Early models like Place2Vec \citep{yan2017itdl}, Doc2Vec~\citep{niu2021delineating} and Location2Vec \citep{qin2022} leveraged spatial co-occurrence among POIs, while graph-based methods \citep{xuyongyang2022, su2023, HGI, huang2022sppe, DCA} modeled functional connectivity among locations.  
These representations highlight the richness of POI data but still offer limited semantic representation capacity. The recent CaLLiPer framework \citep{wang2025multi, liu2025enriching} advances this line of work by aligning POI text embeddings generated by advanced language models, enabling a deeper understanding of urban functions and semantics, although the resulting representations remain limited in capturing high-resolution spatial structure.


Taken together, each data modality offers a unique but incomplete perspective on urban space:  
coordinate data emphasize spatial configuration, EO captures physical characteristics, and POIs encode functional semantics.  
Integrating these complementary signals motivates recent efforts toward multimodal urban representation learning, where contrastive alignment has emerged as a powerful mechanism for unifying heterogeneous sources.

\subsection{Multimodal Alignment and Contrastive Learning}

To overcome the limitations of single-modality representations, recent research increasingly focuses on multimodal alignment frameworks that bridge heterogeneous geospatial signals.
Inspired by CLIP \citep{radford2021clip}, contrastive learning has become a key paradigm for aligning diverse modalities within a shared embedding space, effectively integrating spatial, visual, and semantic information.
In geospatial domains, this idea has been applied to image–location \citep{klemmer2025satclip,vivanco2024geoclip}, image–text \citep{yan2024urbanclip,bai2023geographic}, and POI–location pairings \citep{wang2025multi,liu2025enriching}, demonstrating its potential to unify physical and functional perspectives of space.

Recent multimodal works have expanded this paradigm in different directions.
UrbanCLIP \citep{yan2024urbanclip} employed language-generated captions for satellite imagery to inject textual semantics into visual encoders;
Bai et al. \citep{bai2023geographic}  established an alignment between very-high-resolution satellite imagery and POIs that enables satellite image-based geographic mapping;
MobCLIP \citep{wen2025mobclip} further generalized contrastive alignment by coupling human mobility data with multimodal urban signals, achieving strong cross-domain transfer;
and CaLLiPer \citep{wang2025multi,liu2025enriching} provided a lightweight POI–location alignment framework which produces spatial-semantic embeddings that are effective in geographic mapping and human mobility tasks \citep{wang2025into}.
Together, these studies illustrate the effectiveness of contrastive objectives and projection-based alignment in capturing urban semantics beyond visual appearance.

Nevertheless, existing multimodal frameworks have not explored aligning dense EO foundation model embeddings, such as AlphaEarth, with human-centered semantics.
This motivates our work: AETHER adapts AE to urban contexts through POI-guided multimodal alignment, enriching its EO focused embeddings with functional meaning and advancing toward general-purpose urban representations that couple physical form with functional semantics.

\section{Methodology}
\label{sec:method}

The proposed framework adapts AE embeddings for urban applications through POI-guided contrastive learning. 
Given POIs with textual descriptions and spatial coordinates together with AE embedding maps derived from multi-source EO data, the framework learns a shared embedding space that integrates EO-derived physical features with human-centered semantic signals. 
Specifically, AE provides spatially continuous 64-dimensional representations capturing physical and morphological characteristics of the urban environment, while POIs provide discrete semantic annotations of urban functions. 
By aligning these two modalities, the framework enables AE embeddings to better reflect functional urban semantics while preserving spatial continuity.

\begin{figure}[t]
    \centering
    \includegraphics[width=\linewidth]{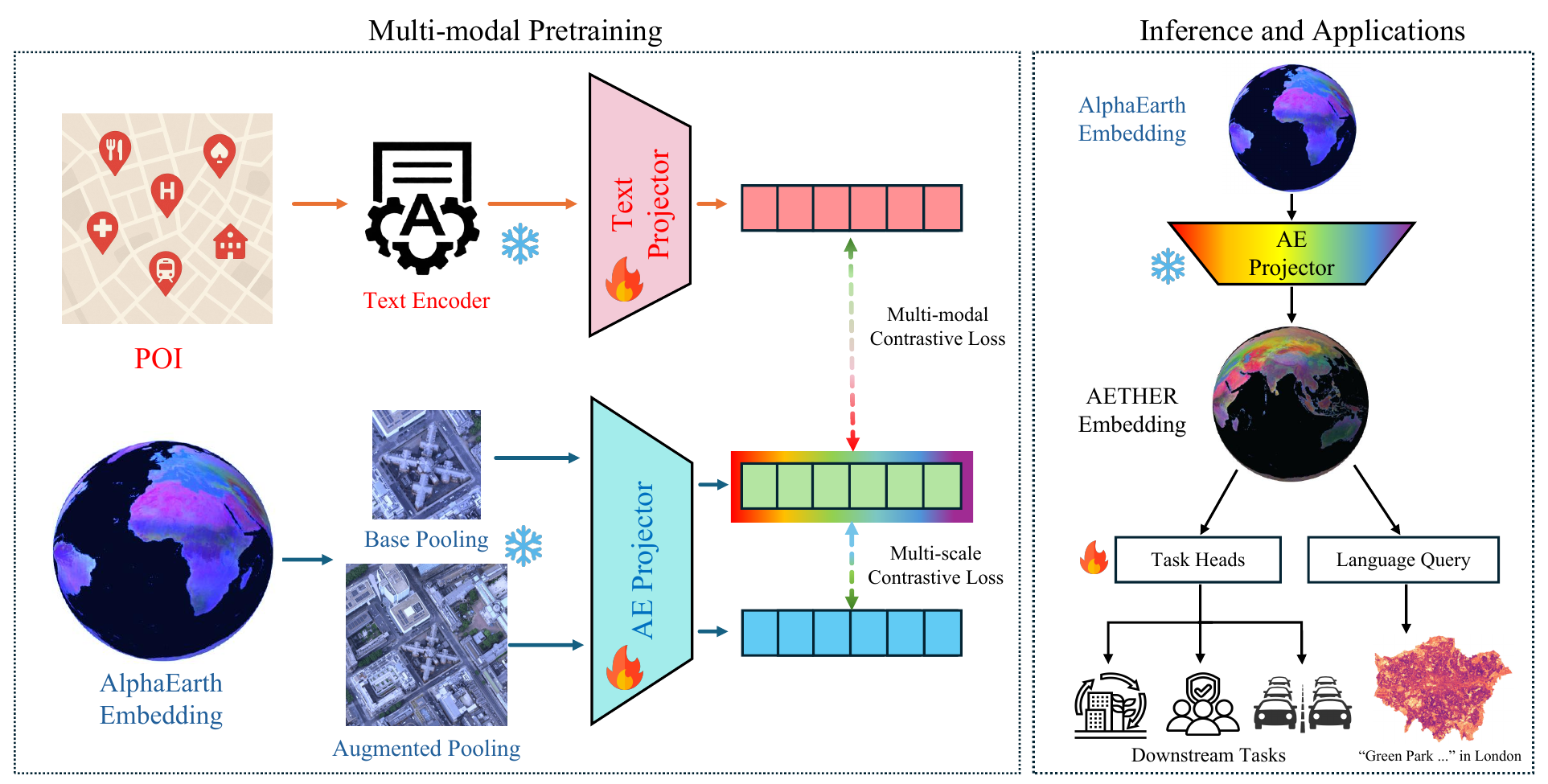}
    \caption{
    Overview of the proposed AETHER framework.
    During multimodal pretraining (left), AE embeddings within dual spatial buffers are pooled and projected, while POI text embeddings are generated via a text encoder and projector.
    Two contrastive losses are applied: a cross-modal AE–POI alignment loss and an intra-modal multi-scale AE--AE consistency loss (weighted by $\lambda$).
    During inference (right), the frozen AE projector produces semantically aligned spatial embeddings, which can be used for downstream tasks or natural language–conditioned spatial retrieval.
    Different colors indicate distinct modalities and representation flows in the multimodal alignment framework.
    }
    \label{fig:framework}
\end{figure}

As shown in Figure \ref{fig:framework}, the AETHER framework consists of three main components:
(1) a POI text branch, which encodes POI descriptions into text embeddings via a pretrained language model and a lightweight projector;
(2) an AE branch, which aggregates local AE embeddings around POI locations using multi-scale pooling and projects them into the same latent space; and
(3) a contrastive alignment module, which jointly optimizes cross-modal (POI–AE) alignment and intra-modal (multi-scale AE) consistency.
During inference, only the AE projector is retained to produce semantically enriched spatial embeddings across the full city, which can be flexibly aggregated for downstream tasks or directly used for natural language–conditioned spatial retrieval.

\subsection{POI Encoder}
\label{sec:method:poi}

POIs are widely used as proxies for human activities and urban functions.  
Recent studies, particularly the CaLLiPer series \citep{wang2025multi, liu2025enriching}, have demonstrated that the rich semantic content encoded in POI names and categories plays a crucial role in constructing meaningful city representations.  
Following this line of work, we employ pretrained language models (LMs) to extract semantic embeddings from POI texts, thereby integrating fine-grained functional meanings into the multimodal representation space shared with AE features.  
The pretrained LM provides contextualized semantics beyond categorical labels, capturing subtle variations and connections in urban functions.

Formally, each POI record \(p_i\) is associated with a text description \(t_i\) that combines both its categorical type and its specific name.  
To capture fine-grained semantics, we construct an enriched description by concatenating the name \(n_i\), first-level category \(c_{1i}\), and second-level subcategory \(c_{2i}\) through a simple template:
\begin{equation}
t_i =``\text{A place of } [c_{2i}], \text{ a type of } [c_{1i}], \text{ named } [n_i]."
\end{equation}
For example, a POI may be represented as ``A place of \textit{coffee shop}, a type of \textit{food and drink}, named \textit{Starbucks}.''  
This formulation allows the model to capture both general functional categories and nuanced semantic distinctions between individual entities.

The pretrained language model \(E(\cdot)\) encodes each description into a contextualized embedding:
\begin{equation}
\mathbf{p}_i = E(t_i) \in \mathbb{R}^{d_t},
\end{equation}
where \(d_t\) denotes the encoder’s output dimension.  
To align with the shared multimodal space, we apply a single linear projector and $\ell_2$ normalization:
\begin{equation}
\mathbf{z}_i^{\text{poi}} = \mathrm{Norm}(W_{\text{poi}}\mathbf{p}_i),
\qquad
W_{\text{poi}} \in \mathbb{R}^{d\times d_t}.
\end{equation}
This enriched text representation enables the model to encode both functional and entity-level semantics, 
improving its ability to differentiate between places with similar categories but distinct social or economic connotations.

\subsection{AlphaEarth Encoder}
\label{sec:method:ae}

AE \citep{brown2025alphaearth} provides high-quality geospatial embeddings that integrate multi-source EO data through a unified space-time precision encoder.  
Each 10\,m pixel on Earth's surface is represented by a 64-dimensional embedding that summarizes local spectral, temporal, and environmental contexts within a defined temporal window.  
These embeddings are released on Google Earth Engine, forming a globally consistent and information-rich base for downstream mapping applications.

For each POI, we retrieve AE features from the embedding field at or around its coordinates. 
To incorporate local spatial context and mitigate local noise, we extract features within spatial windows centered around each POI location. 
Specifically, we define a \emph{base view} with a square window of radius \(r_b\) and an \emph{augmented view} with a larger radius \(r_a > r_b\). 
Average pooling is applied within each window to obtain two spatially aggregated feature vectors, \emph{i.e.}, $\mathbf{a}_i^{(b)},\ \mathbf{a}_i^{(a)} \in \mathbb{R}^{64}$. 
This multi-scale design mitigates the mismatch between point-based POI annotations and the spatial extent of real-world urban objects. 
By aggregating AE embeddings across multiple spatial scales, the model captures broader spatial context and learns scale-consistent representations that are more robust to local variability.

The aggregated AE features are then passed through a lightweight projection head \(f_\theta(\cdot)\), implemented as a two-layer MLP with residual and gating connections followed by a post-MLP projection, and $\ell_2$ normalization:
\begin{equation}
\mathbf{z}_i^{(b)} = 
\frac{f_\theta(\mathbf{a}_i^{(b)})}{\|f_\theta(\mathbf{a}_i^{(b)})\|_2}, 
\qquad 
\mathbf{z}_i^{(a)} = 
\frac{f_\theta(\mathbf{a}_i^{(a)})}{\|f_\theta(\mathbf{a}_i^{(a)})\|_2}.
\end{equation}

The outputs \(\mathbf{z}_i^{(b)}\) and \(\mathbf{z}_i^{(a)}\) correspond to normalized embeddings from the base and augmented spatial views, respectively, both lying on the unit hypersphere \(\mathbb{S}^{d-1}\).  
This projection step adapts the 64-dimensional AE features into the shared latent space (\(d\)-dimensional) used for multimodal alignment with POI semantics.

This design preserves AE’s physical priors while enabling semantic alignment.

\subsection{Multimodal Contrastive Alignment}
\label{sec:method:alignment}

We align the AE and POI representations in a shared latent space through contrastive learning.  
For a batch of \(N\) POIs, we obtain three embeddings \((\mathbf{z}_i^{(b)}, \mathbf{z}_i^{(a)}, \mathbf{z}_i^{\text{poi}})\),  
where \(\mathbf{z}_i^{(b)}\) and \(\mathbf{z}_i^{(a)}\) denote the base and augmented AE embeddings, and \(\mathbf{z}_i^{\text{poi}}\) is the POI embedding.  
The alignment objective consists of two symmetric contrastive losses:  
an intra-modal AE--AE consistency term and a cross-modal AE–POI alignment term.
The base-view embeddings are used for cross-modal alignment with POI semantics,
while the augmented view provides a complementary spatial context for intra-modal consistency.

\paragraph{(1) Intra-modal AE--AE consistency}
This term enforces representation stability across different spatial scales of the same location,  
encouraging the AE encoder to produce consistent embeddings for nearby contexts:
\begin{equation}
\label{eq:l_ii}
\mathcal{L}_{\mathrm{AA}} =
-\frac{1}{2N} \sum_{i=1}^{N} 
\bigg[
\log \frac{\exp(\mathbf{z}_i^{(b)\top}\mathbf{z}_i^{(a)}/\tau_{\mathrm{ae}})}
{\sum_{j=1}^{N}\exp(\mathbf{z}_i^{(b)\top}\mathbf{z}_j^{(a)}/\tau_{\mathrm{ae}})}
+
\log \frac{\exp(\mathbf{z}_i^{(a)\top}\mathbf{z}_i^{(b)}/\tau_{\mathrm{ae}})}
{\sum_{j=1}^{N}\exp(\mathbf{z}_i^{(a)\top}\mathbf{z}_j^{(b)}/\tau_{\mathrm{ae}})}
\bigg],
\end{equation}
where \(\tau_{\mathrm{ae}}\) is a fixed temperature controlling the sharpness of the similarity distribution.

\paragraph{(2) Cross-modal AE–POI alignment}
This term introduces human-centered semantic supervision by aligning AE embeddings with POI representations in the shared latent space:
\begin{equation}
\label{eq:l_xt}
\mathcal{L}_{\mathrm{AP}} =
-\frac{1}{2N} \sum_{i=1}^{N} 
\bigg[
\log \frac{\exp(\mathbf{z}_i^{(b)\top}\mathbf{z}_i^{\text{poi}}/\tau_{\mathrm{poi}})}
{\sum_{j=1}^{N}\exp(\mathbf{z}_i^{(b)\top}\mathbf{z}_j^{\text{poi}}/\tau_{\mathrm{poi}})}
+
\log \frac{\exp(\mathbf{z}_i^{\text{poi}\top}\mathbf{z}_i^{(b)}/\tau_{\mathrm{poi}})}
{\sum_{j=1}^{N}\exp(\mathbf{z}_i^{\text{poi}\top}\mathbf{z}_j^{(b)}/\tau_{\mathrm{poi}})}
\bigg],
\end{equation}
where \(\tau_{\mathrm{poi}}\) denotes the temperature for cross-modal alignment.

\paragraph{(3) Overall objective}
The final training loss combines the two objectives with a balancing coefficient \(\lambda \in [0,1)\):
\begin{equation}
\label{eq:total_loss}
\mathcal{L} = \lambda\,\mathcal{L}_{\mathrm{AA}} + (1 - \lambda)\,\mathcal{L}_{\mathrm{AP}}.
\end{equation}
This joint optimization encourages spatial consistency while aligning embeddings with POI semantics.
The model is optimized using AdamW with fixed temperatures.

\subsection{Downstream Tasks}
\label{sec:method:downstream}

After multimodal alignment, the trained AE projector $f_\theta$
is applied across the embedding field to generate
semantically enriched spatial representations.
For a spatial unit $k$ (\emph{e.g.}, point buffer, administrative polygon, H3 cell, or raster grid),
we aggregate base-view embeddings by mean pooling:
\begin{equation}
\mathbf{r}_k = \frac{1}{N_k} \sum_{i \in \text{region }k} \mathbf{z}_i^{(b)} \in \mathbb{R}^{d}.
\end{equation}

A lightweight task head $h_\psi(\cdot)$ is attached to $\mathbf{r}_k$.
In our experiments, $h_\psi$ is implemented as a two-layer MLP.

We evaluate the learned representations on four urban tasks
across two cities, spanning discrete classification,
compositional distribution modeling, and scalar regression.

In London, land-use classification (LUC) predicts the land-use category of sampled points, with discrete labels derived from the NLUD dataset and optimized using a standard cross-entropy objective. 
Socioeconomic distribution mapping (SDM) operates at the LSOA level, where the model predicts a normalized occupational distribution; training minimizes distributional cross-entropy.

In Singapore, land-use distribution mapping (LUD) predicts proportional land-use distributions defined over H3 hexagonal cells, following the same distributional formulation as SDM. 
GDP mapping is formulated as scalar regression, where each spatial unit is associated with a log-standardized GDP per capita value and optimized using mean squared error.

\paragraph{Training protocol}

For all tasks, region embeddings $\{\mathbf{r}_k\}$ are fixed,
and only the task head parameters $\psi$ are optimized.
Models are trained using Adam with early stopping on validation loss.
Random train/validation/test splits are repeated across five seeds,
and all baselines follow the same protocol for fair comparison.

\subsection{Natural Language–Conditioned Spatial Retrieval}
\label{sec:method:retrieval}

In addition to supervised downstream tasks, the aligned embedding enables direct natural language–conditioned spatial retrieval.
Given a free-form textual query $q$, we encode it using the same
pretrained language model $E(\cdot)$ and apply the learned text projector,
followed by $\ell_2$ normalization:
\begin{equation}
\mathbf{z}_q^{\text{text}} =
\mathrm{Norm}\!\left(W_{\text{poi}} E(q)\right)
\in \mathbb{R}^{d}.
\end{equation}

Let $\mathbf{z}_i \in \mathbb{S}^{d-1}$ denote the normalized AETHER embedding
at spatial location $i$, obtained from the aligned AE projector.
We compute cosine similarity across the spatial domain:
\begin{equation}
s(i \mid q) =
\mathbf{z}_q^{\text{text}\top}
\mathbf{z}_i.
\end{equation}

The similarity field $s(i \mid q)$ defines a semantic response map,
indicating the degree of alignment between the textual query
and geographic locations.

We consider two complementary evaluation settings:

\paragraph{(1) Qualitative spatial analysis}
The similarity map is visualized and compared with authoritative
ground-truth datasets to assess whether high-response regions
correspond to semantically relevant spatial structures.

\paragraph{(2) Quantitative retrieval evaluation}
Spatial units are ranked by similarity score $s(i \mid q)$.
For a given threshold $k\%$, the top-$k\%$ highest-scoring units
are selected.
Recall@Top-$k\%$ is defined as:
\begin{equation}
\text{Recall@Top-}k\% =
\frac{
\left| \mathcal{P}_{\text{retrieved}}^{(k)} \right|
}{
\left| \mathcal{P}_{\text{gt}} \right|
},
\end{equation}
where $\mathcal{P}_{\text{gt}}$ denotes the set of ground-truth POI instances,
and $\mathcal{P}_{\text{retrieved}}^{(k)}$ denotes the subset
whose corresponding spatial locations fall within the top-$k\%$
similarity region.

\section{Experimental Setup}
\label{sec:exp}

\subsection{Study Areas and Datasets}
\label{sec:exp:data}

Experiments are conducted in two metropolitan regions, Greater London (UK) and Singapore, 
to evaluate geographical robustness under different urban morphologies and climatic contexts. 
For both cities, we use the publicly released AE embeddings with 10 m resolution (Section~\ref{sec:method:ae}), serving as the frozen EO backbone throughout all experiments.

\paragraph{Data sources}
\begin{itemize}
    \item \textbf{AlphaEarth embeddings}\footnote{\url{https://earthengine.google.com/}} \citep{brown2025alphaearth}. 
    Annualized 64-dimensional embedding fields derived from multi-source Earth Observation data.

    \item \textbf{POIs}. 
    For London, we use Ordnance Survey POI data\footnote{\url{https://digimap.edina.ac.uk/}}, 
    containing geographic coordinates, names, and hierarchical categories. 
    For Singapore, we use Foursquare POI data \footnote{\url{https://foursquare.com/products/places/}} with comparable metadata fields.

    \item \textbf{Land-use data}. 
    For London, the Verisk National Land Use Database (NLUD)\footnote{\url{https://digimap.edina.ac.uk/roam/map/verisk}} 
    provides polygon-based land-use labels across ten major categories. 
    For Singapore, the URA Master Plan 2019 land-use layer\footnote{\url{https://data.gov.sg/dataset/master-plan-2019-land-use-layer}} 
    provides statutory land-use designations.

    \item \textbf{Socioeconomic data}. 
    For London, we use the 2021 ONS Census National Statistics Socioeconomic Classification (NS-SeC) 
    distributions at the LSOA level\footnote{\url{https://www.ons.gov.uk/}}. 
    For Singapore, we use the downscaled gridded global GDP per capita (PPP) dataset (2020)\footnote{\url{https://zenodo.org/records/16741980}} 
    \citep{kummu2025downscaled}.
\end{itemize}

All data is projected to the respective local coordinate systems (London: EPSG:27700; Singapore: EPSG:3414), clipped to city boundaries, 
and spatially aligned with AE embeddings prior to feature extraction. 
POIs are cleaned and standardized into unified textual descriptions by combining the venue name, primary category, and fine-grained subcategory for language-model-based text embedding.

\subsection{Pretraining Setup}

The multimodal alignment between AE and POI embeddings follows the formulation in Section~\ref{sec:method:alignment}.  
For each POI, two spatial views are constructed by aggregating AE embeddings 
within fixed-size square windows centered at the corresponding raster location.  
Given a pixel radius \(r\), embeddings are mean-pooled over a 
\((2r+1)\times(2r+1)\) neighborhood in raster space.
In practice, the \textit{base view} uses a radius \(r_b = 4\), 
while the \textit{augmented view} uses a larger radius \(r_a = 9\).  
These radii correspond to spatial windows of approximately \(9\times9\) and \(19\times19\) pixels.
POI textual descriptions are encoded using the pretrained 
\texttt{text-embedding-3-large} model from OpenAI,
with the output embedding dimension set to 384 for efficiency.

The total loss is defined as
\(\mathcal{L}=\lambda\,\mathcal{L}_{\mathrm{AA}} + (1-\lambda)\,\mathcal{L}_{\mathrm{AP}}\),
where \(\lambda\) balances the two objectives.
In our experiments, we set \(\lambda=0.2\) and use fixed temperatures
\(\tau_{\mathrm{ae}}=\tau_{\mathrm{poi}}=0.07\).
The influence of \(\lambda\) is further analyzed in the sensitivity study.
We optimize the model using AdamW with a batch size of 512.  
The hidden dimension is set to 256 and the output embedding dimension to 128.
Training is performed for 100 epochs on a single NVIDIA RTX PRO 6000 (96\,GB) GPU.  
The last checkpoint of epoch 100 is selected.

\subsection{Analysis Units and Spatial Aggregation}

Experiments are conducted in London and Singapore 
across discrete, compositional, and continuous spatial mapping tasks.

\paragraph{London}
For LUC, point samples are derived from NLUD polygons, 
each assigned a discrete land-use category. 
For downstream evaluation, AE embeddings are aggregated within a 50\,m circular buffer 
centered at each sample location.
For SDM, analysis is performed at the LSOA level.  
Each LSOA polygon is associated with a normalized NS-SeC occupational distribution.  
Region-level representations are obtained by averaging AE embeddings 
over all pixels contained within each administrative unit.

\paragraph{Singapore}
For LUD, spatial units are defined 
using an H3 hexagonal grid at resolution 9 
(average cell area $\approx$ 0.1 km$^2$, edge length $\approx$ 170 m). 
Master Plan 2019 land-use polygons are intersected with H3 cells 
to compute proportional land-use compositions within each hexagon. 
AE embeddings are aggregated over pixels contained within each H3 cell.
For GDP mapping, a gridded GDP per capita (PPP) dataset 
with 1 km spatial resolution is used.  
Each valid raster cell is treated as an independent spatial unit, 
and AE embeddings are aggregated within a 500\,m circular buffer 
centered at the grid-cell location.

\subsection{Baselines}
\label{sec:exp:baselines}

We compare AETHER with representative baselines spanning random, semantic, spatial, POI-driven and EO-driven urban representation models:

\begin{itemize}
    \item Random baseline.
    Randomly initialized embeddings with the same dimensionality as AETHER, 
    serving as a lower-bound reference.

    \item Simple Semantic baseline.
    LDA \citep{blei2003latent} constructs region-level topic representations 
    based on POI category co-occurrence, providing an interpretable 
    text-driven urban representation.

    \item Coordinate-based spatial encoder.
    Space2Vec \citep{mai2020iclr} learns continuous spatial representations 
    directly from geographic coordinates.

    \item POI-based models. 
    Place2Vec \citep{yan2017itdl}, Doc2Vec \citep{niu2021delineating}, and
    SPPE \citep{huang2022sppe} learn urban representations 
    from POI distributions, spatial co-occurrences, or graph-structured relations.

    \item Multimodal POI–location alignment.
    CaLLiPer \citep{wang2025multi, liu2025enriching} aligns POI text embeddings 
    with spatial coordinates via contrastive learning.

    \item Earth-observation models.
    SatCLIP \citep{klemmer2025satclip} and AlphaEarth \citep{brown2025alphaearth} 
    produce embeddings trained on Earth Observation data. 
    AlphaEarth provides 64-dimensional embeddings at 10\,m resolution 
    and serves as the EO backbone in our framework.
\end{itemize}

\subsection{Evaluation Metrics}
\label{sec:exp:metrics}

We evaluate AETHER using task-specific metrics.  
All results are reported as mean $\pm$ standard deviation over five random seeds.  
For all metrics, $\uparrow$ indicates that higher values denote better performance, 
while $\downarrow$ indicates the opposite.

\paragraph{Land-Use Classification (LUC)}
LUC is formulated as a multi-class classification task.  
We report macro-averaged F1 score, precision, and recall.  
Macro-averaging computes the unweighted mean across all classes, 
ensuring equal importance for minority categories.  

\paragraph{Distribution Mapping (SDM and LUD)}
For SDM in London and LUD in Singapore, 
the target is a normalized probability distribution 
$\mathbf{q} \in \Delta^{C-1}$ for each spatial unit.  
We evaluate predicted distributions using:

\begin{itemize}[leftmargin=1.5em]
    \item Kullback–Leibler (KL) divergence ($\downarrow$), quantifying information divergence between predicted and true distributions.
    \item $L_1$ distance ($\downarrow$), measuring the mean absolute difference between predicted and ground-truth distributions;
    \item Chebyshev distance ($\downarrow$), capturing the maximum absolute deviation across distribution components;
\end{itemize}

\paragraph{GDP Mapping}
GDP prediction is formulated as a continuous regression task.  
We report the coefficient of determination ($R^2$, $\uparrow$), 
mean absolute error (MAE, $\downarrow$), and root mean squared error (RMSE, $\downarrow$).  

\subsection{Natural Language–Conditioned Spatial Retrieval}
\label{sec:exp:retrieval}
\paragraph{Query formulation}
Textual queries are constructed in two forms.
For qualitative analysis, we use free-form functional phrases 
(\emph{e.g.}, ``Airport and Transportation Hub'', 
``Central Business District'', 
``Green Park and Natural Reserve'') 
to assess open-vocabulary localization ability across cities.
For quantitative evaluation, we generate category-level prompts 
based on POI class names using the template:
``A place with the type of [category].''
All queries are encoded using the same pretrained language model 
and projected via the trained text projector described in Section~\ref{sec:method:poi}.

\paragraph{Candidate space}
Retrieval is performed over all spatial embeddings 
generated by the aligned AE projector within each city.
Cosine similarity is computed between the normalized query embedding 
and every spatial embedding to produce a full-coverage similarity field.

\paragraph{Qualitative evaluation}
Qualitative retrieval is conducted in both London and Singapore.
Full-city similarity maps are visualized and compared with authoritative 
planning layers and public geospatial datasets to assess semantic 
localization consistency and cross-city generalization.

\paragraph{Quantitative evaluation}
Quantitative retrieval is conducted in Greater London, 
which serves as the primary benchmark city.
Compared with Singapore, London covers a larger metropolitan extent,
exhibits greater functional heterogeneity, and provides a more 
internally consistent POI taxonomy, enabling controlled 
category-level evaluation.

For selected mid-frequency POI classes 
(\emph{e.g.}, Art and Antiques, Religious Organisations, 
Medical Equipment, Supplies and Pharmaceuticals, Sports Clubs and Associations, Import and Export Services),
spatial pixels are ranked by cosine similarity.
Recall@Top-$k\%$ is computed for $k \in \{1,5,10\}$.
A POI instance is considered successfully retrieved 
if the spatial pixel containing the POI falls within the top-$k\%$ 
highest-scoring pixels.
We compare AETHER with three baselines: AlphaEarth
\citep{brown2025alphaearth}, Random, and CaLLiPer 
\citep{wang2025multi, liu2025enriching}.

\subsection{Ablation Study and Sensitivity Analysis}
\label{sec:exp:sensitivity}

We evaluate the robustness of AETHER with respect to two key factors:
the loss balance coefficient $\lambda$ and the amount of training supervision.
All experiments are conducted in both London and Singapore,
varying one factor at a time while keeping the other fixed at the canonical configuration
($\lambda=0.2$, full training set).
All evaluations follow the protocol described in Section~\ref{sec:exp:metrics}.

\paragraph{Loss balance ($\lambda$)}
We examine the effect of the contrastive weighting coefficient
$\lambda \in \{0.0, 0.2, 0.4, 0.6, 0.8\}$
in the overall objective
$\mathcal{L}=\lambda\,\mathcal{L}_{\mathrm{AA}}+(1-\lambda)\,\mathcal{L}_{\mathrm{AP}}$.
This analysis investigates how the trade-off between
intra-modal AE--AE consistency and cross-modal AE–POI alignment
affects downstream performance.

\paragraph{Training supervision}
To assess data efficiency, we subsample the POI–AE pairs
to proportions $\{0.2, 0.4, 0.6, 0.8, 1\}$ of the full training set, sampled uniformly at random.
This experiment evaluates how performance scales with supervision strength and whether the alignment mechanism remains stable under reduced training data.

These ablation experiments analyze the sensitivity of AETHER
to objective weighting and supervision scale across two urban contexts.
Quantitative results are reported in Section~\ref{sec:results}.

\section{Results}
\label{sec:results}

\subsection{Downstream Performance}
\label{sec:results:downstream}

Tables~\ref{tab:london_results} and~\ref{tab:singapore_results}
report results across four downstream tasks in Greater London and Singapore.
Across both cities and all four tasks, AETHER achieves the best performance among all baselines, with relative improvements over the best baselines ranging from 4.5\% to 21.9\%.
This demonstrates that POI-guided alignment
systematically enhances the representational capacity
of AlphaEarth embeddings.

\paragraph{City-specific task characteristics}
The two study areas exhibit distinct spatial configurations that influence how much task-relevant information is in the embeddings from different data sources.

Greater London spans a large metropolitan region
with extensive natural land, suburban neighborhoods,
and heterogeneous built forms.
In this context, LUC is primarily driven by morphological and land-cover cues that are effectively captured by EO embeddings.
Accordingly, AlphaEarth already performs strongly on LUC, substantially outperforming POI-trained models,
reflecting its ability to encode structural and surface characteristics.
In contrast, SDM requires inference of demographic
and functional attributes that are only weakly reflected in physical appearance.
On this task, AlphaEarth exhibits relatively modest performance, while semantically informed POI-based models, particularly CaLLiPer, provide the strongest baseline, highlighting the importance of functional semantics.

Singapore presents a denser and more functionally mixed environment, where residential, commercial, and infrastructural uses are highly interwoven at fine spatial scales. 
In such compact and vertically structured settings, purely morphological cues derived from EO data are less sufficient for disentangling functional composition. 
Accordingly, tasks such as LUD and GDP prediction rely more heavily on latent functional semantics related to human activities and urban functions.
For LUD, AlphaEarth still outperforms most baseline models, indicating that morphological signals remain informative. 
However, for GDP prediction, several POI-based models (\emph{e.g.}, LDA, SPPE, and CaLLiPer) surpass AlphaEarth, suggesting that socioeconomic attributes are better captured through human-centered semantic supervision than through spectral features alone.

\begin{table*}[!ht]
\caption{
Experimental results in London.  
All metrics are reported as mean $\pm$ std over five random seeds.  
For LUC, higher values indicate better performance; for SDM, lower values are better.  
For readability, metrics with small magnitudes are rescaled for presentation:
all LUC metrics (F1, precision, recall) are multiplied by $10^{2}$,
while for SDM, $L_{1}$ distances are multiplied by $10^{2}$ and both KL divergence and Chebyshev distances are multiplied by $10^{3}$.
Best and second-best results are shown in \textbf{bold} and \underline{underlined}, respectively.
The last row (Improv.) reports the relative improvement of AETHER
with respect to the strongest baseline for each metric.
}
\label{tab:london_results}
\centering
\small
\setlength{\tabcolsep}{6pt}
\begin{tabular}{l|ccc|ccc}
\toprule
\multirow{2}{*}{Model} &
\multicolumn{3}{c|}{LUC (↑)} &
\multicolumn{3}{c}{SDM (↓)} \\
\cmidrule(lr){2-4}\cmidrule(lr){5-7}
& F1 & Precision & Recall &
KL & L1 & Chebyshev \\
\midrule
Random & 6.6 $\pm$ 1.8 & 8.3 $\pm$ 1.4 & 10.3 $\pm$ 0.6 & 75.7 $\pm$ 1.1 & 30.0 $\pm$ 0.2 & 91.4 $\pm$ 0.8 \\
LDA & 33.0 $\pm$ 1.3 & 31.5 $\pm$ 1.2 & 37.3 $\pm$ 1.8 & 50.7 $\pm$ 0.9 & 23.8 $\pm$ 0.2 & 71.8 $\pm$ 0.8 \\
Place2Vec & 33.1 $\pm$ 2.1 & 33.3 $\pm$ 2.1 & 36.4 $\pm$ 2.6 & 48.0 $\pm$ 0.9 & 23.0 $\pm$ 0.3 & 69.2 $\pm$ 0.6 \\
Doc2Vec & 32.7 $\pm$ 0.6 & 34.9 $\pm$ 0.9 & 33.8 $\pm$ 0.5 & 47.5 $\pm$ 0.8 & 23.1 $\pm$ 0.2 & 68.9 $\pm$ 0.6 \\
SPPE & 34.1 $\pm$ 1.3 & 34.1 $\pm$ 1.3 & 37.8 $\pm$ 1.4 & 45.0 $\pm$ 0.8 & 22.3 $\pm$ 0.2 & 67.4 $\pm$ 0.7 \\
Space2Vec & 26.0 $\pm$ 1.2 & 28.2 $\pm$ 2.0 & 30.3 $\pm$ 1.0 & 46.4 $\pm$ 1.3 & 22.7 $\pm$ 0.3 & 68.6 $\pm$ 1.0 \\
CaLLiPer & 35.5 $\pm$ 0.6 & 34.3 $\pm$ 0.5 & 39.3 $\pm$ 0.7 & \underline{34.1 $\pm$ 1.3} & \underline{19.3 $\pm$ 0.4} & \underline{58.0 $\pm$ 1.4} \\
SatCLIP & 12.2 $\pm$ 1.6 & 12.2 $\pm$ 1.5 & 16.8 $\pm$ 2.0 & 70.7 $\pm$ 0.1 & 28.9 $\pm$ 0.1 & 88.4 $\pm$ 0.3 \\
AlphaEarth & \underline{55.9 $\pm$ 1.5} & \underline{55.8 $\pm$ 1.4} & \underline{57.4 $\pm$ 1.7} & 42.2 $\pm$ 1.1 & 21.5 $\pm$ 0.4 & 64.8 $\pm$ 0.9 \\
\midrule
AETHER (ours) & \textbf{59.4 $\pm$ 0.9} & \textbf{58.3 $\pm$ 1.0} & \textbf{61.8 $\pm$ 0.9} & \textbf{30.1 $\pm$ 1.2} & \textbf{17.8 $\pm$ 0.3} & \textbf{53.8 $\pm$ 1.1} \\
Improv. 
& \textcolor{DarkGreen}{6.3\%} 
& \textcolor{DarkGreen}{4.5\%} 
& \textcolor{DarkGreen}{7.7\%} 
& \textcolor{DarkGreen}{11.7\%} 
& \textcolor{DarkGreen}{7.8\%} 
& \textcolor{DarkGreen}{7.2\%} \\
\bottomrule
\end{tabular}
\end{table*}

\begin{table*}[!ht]
\caption{
Experimental results in Singapore.
All metrics are reported as mean $\pm$ std over five random seeds.
For LUD and GDP, lower values indicate better performance, except for $R^2$, where higher values indicate better performance.
For readability, metrics with small magnitudes are rescaled for presentation:
all LUD metrics (KL, $L_1$, and Chebyshev) are multiplied by $10^{2}$,
while GDP $R^2$ is expressed in percentage (\%).
Best and second-best results are shown in \textbf{bold} and \underline{underlined}, respectively.
The last row (Improv.) reports the relative improvement of AETHER
with respect to the strongest baseline for each metric.
}
\label{tab:singapore_results}
\centering
\small
\setlength{\tabcolsep}{6pt}
\begin{tabular}{l|ccc|ccc}
\toprule
\multirow{2}{*}{Model} &
\multicolumn{3}{c|}{LUD (↓)} &
\multicolumn{3}{c}{GDP} \\
\cmidrule(lr){2-4}\cmidrule(lr){5-7}
& KL & L1 & Chebyshev &
R2 (↑) & MAE (↓) & RMSE (↓) \\
\midrule
Random & 147.8 $\pm$ 0.8 & 141.1 $\pm$ 0.3 & 62.3 $\pm$ 0.3 & 8.4 $\pm$ 2.6 & 88.2 $\pm$ 1.6 & 95.3 $\pm$ 0.7 \\
LDA & 102.4 $\pm$ 0.7 & 104.8 $\pm$ 0.6 & 47.5 $\pm$ 0.3 & 73.6 $\pm$ 1.9 & 33.8 $\pm$ 1.3 & 51.2 $\pm$ 1.5 \\
Place2Vec & 104.4 $\pm$ 1.2 & 105.6 $\pm$ 0.6 & 47.7 $\pm$ 0.3 & 65.2 $\pm$ 4.2 & 36.5 $\pm$ 1.9 & 58.7 $\pm$ 3.3 \\
Doc2Vec & 108.7 $\pm$ 1.6 & 111.1 $\pm$ 0.8 & 50.7 $\pm$ 0.4 & 63.1 $\pm$ 3.5 & 42.1 $\pm$ 2.2 & 60.5 $\pm$ 2.5 \\
SPPE & 99.8 $\pm$ 1.1 & 103.0 $\pm$ 0.6 & 46.9 $\pm$ 0.3 & 72.7 $\pm$ 2.2 & \underline{32.2 $\pm$ 1.0} & 52.0 $\pm$ 1.5 \\
Space2Vec & 85.6 $\pm$ 1.5 & 92.6 $\pm$ 1.1 & 41.5 $\pm$ 0.5 & 65.1 $\pm$ 3.1 & 41.4 $\pm$ 2.1 & 58.8 $\pm$ 1.9 \\
CaLLiPer & \underline{54.9 $\pm$ 1.5} & \underline{65.1 $\pm$ 1.1} & \underline{28.8 $\pm$ 0.5} & \underline{76.2 $\pm$ 1.6} & 33.4 $\pm$ 1.0 & \underline{48.5 $\pm$ 1.3} \\
SatCLIP & 111.4 $\pm$ 0.9 & 117.8 $\pm$ 0.3 & 52.1 $\pm$ 0.3 & 31.7 $\pm$ 4.5 & 69.2 $\pm$ 1.9 & 82.2 $\pm$ 1.9 \\
AlphaEarth & 64.0 $\pm$ 2.8 & 73.9 $\pm$ 1.9 & 32.6 $\pm$ 1.0 & 70.6 $\pm$ 0.3 & 37.1 $\pm$ 0.6 & 54.0 $\pm$ 0.4 \\
\midrule
AETHER & \textbf{42.9 $\pm$ 1.6} & \textbf{53.3 $\pm$ 1.6} & \textbf{23.5 $\pm$ 0.7} & \textbf{80.9 $\pm$ 2.6} & \textbf{28.1 $\pm$ 1.8} & \textbf{43.4 $\pm$ 3.1} \\
Improv. & \textcolor{DarkGreen}{21.9\%} & \textcolor{DarkGreen}{18.1\%} & \textcolor{DarkGreen}{18.4\%} & \textcolor{DarkGreen}{6.2\%} & \textcolor{DarkGreen}{12.7\%} & \textcolor{DarkGreen}{10.5\%} \\

\bottomrule
\end{tabular}
\end{table*}

\paragraph{Variation in improvement magnitude}
Consistent with the task characteristics discussed above,
the magnitude of improvement brought by AETHER varies across tasks, reflecting how much functional semantics are missing from the original, unaligned EO embeddings.
In London LUC, where morphological signals already provide
strong discriminative power and AlphaEarth performs competitively, AETHER still delivers a moderate but consistent gain (+6.3\% macro-F1 over AE).
This indicates that POI semantic alignment enhances class separability even in settings where EO features are already strong, providing complementary functional cues rather than replacing morphological information.

By contrast, for socially grounded tasks such as London SDM and Singapore GDP prediction, performance gains are substantially larger.
For SDM, KL divergence is reduced by 28.7\% relative to AE,
and for LUD prediction, AETHER achieves the highest $R^2$ (80.9\%) with notable reductions in MAE and RMSE.
These larger improvements suggest that cross-modal alignment
compensates for semantic information that is largely absent
from the original AlphaEarth embeddings.
The results therefore demonstrate both the necessity and effectiveness of injecting urban functional semantics, leading to a more pronounced restructuring of the embedding space toward functionally meaningful organization.

Overall, the consistent state-of-the-art results across heterogeneous urban contexts support the hypothesis that
POI-guided contrastive alignment systematically enriches foundation-model embeddings, improving both morphologically and semantically driven tasks while preserving spatial coherence.

\subsection{Spatial Embedding Visualization}
\label{sec:results:mechanism}

\begin{figure*}[t]
\centering

\begin{subfigure}[t]{0.23\linewidth}
    \centering
    \includegraphics[width=\linewidth]{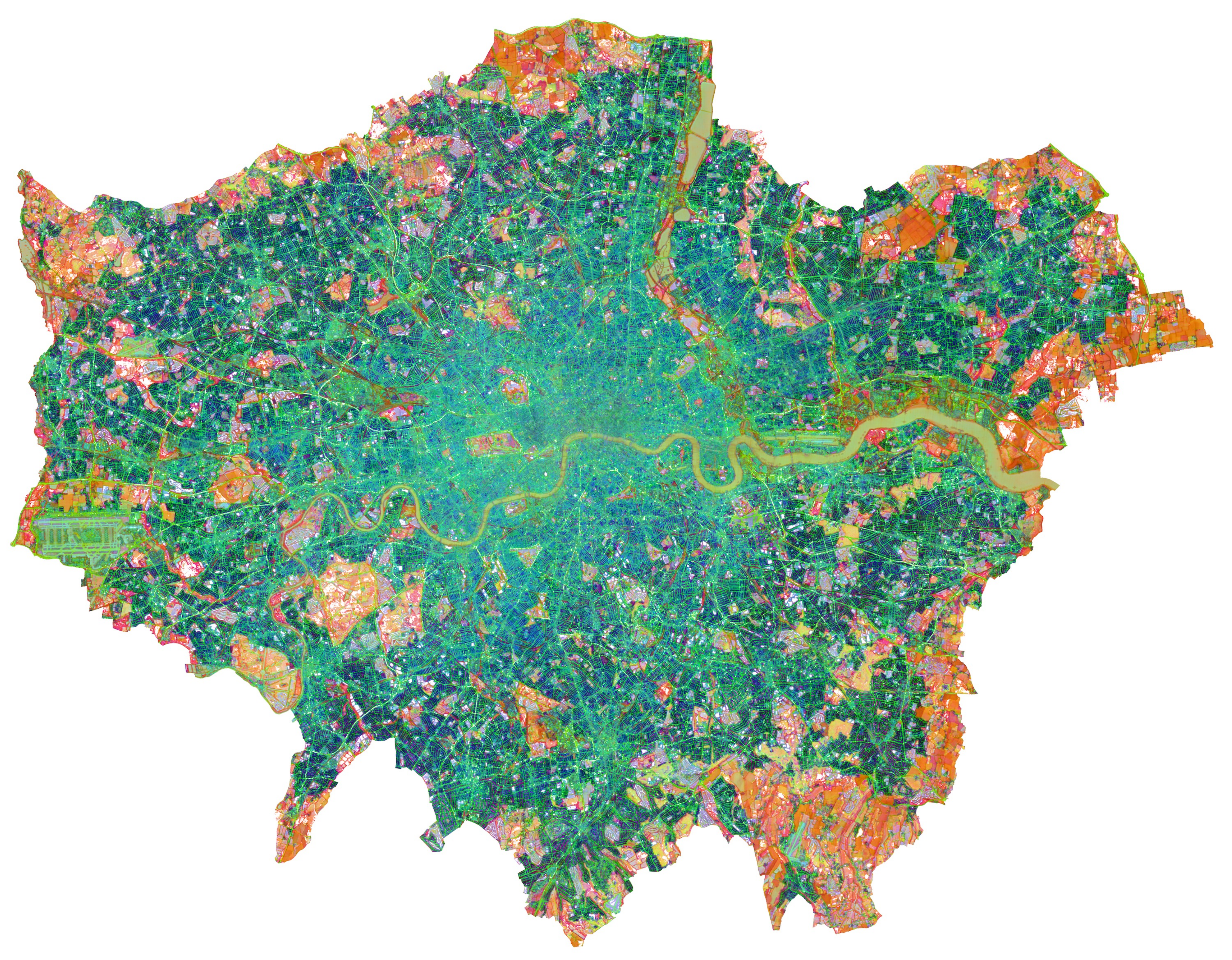}
    \caption{London -- AETHER}
    \label{fig:ld_aether}
\end{subfigure}
\hfill
\begin{subfigure}[t]{0.23\linewidth}
    \centering
    \includegraphics[width=\linewidth]{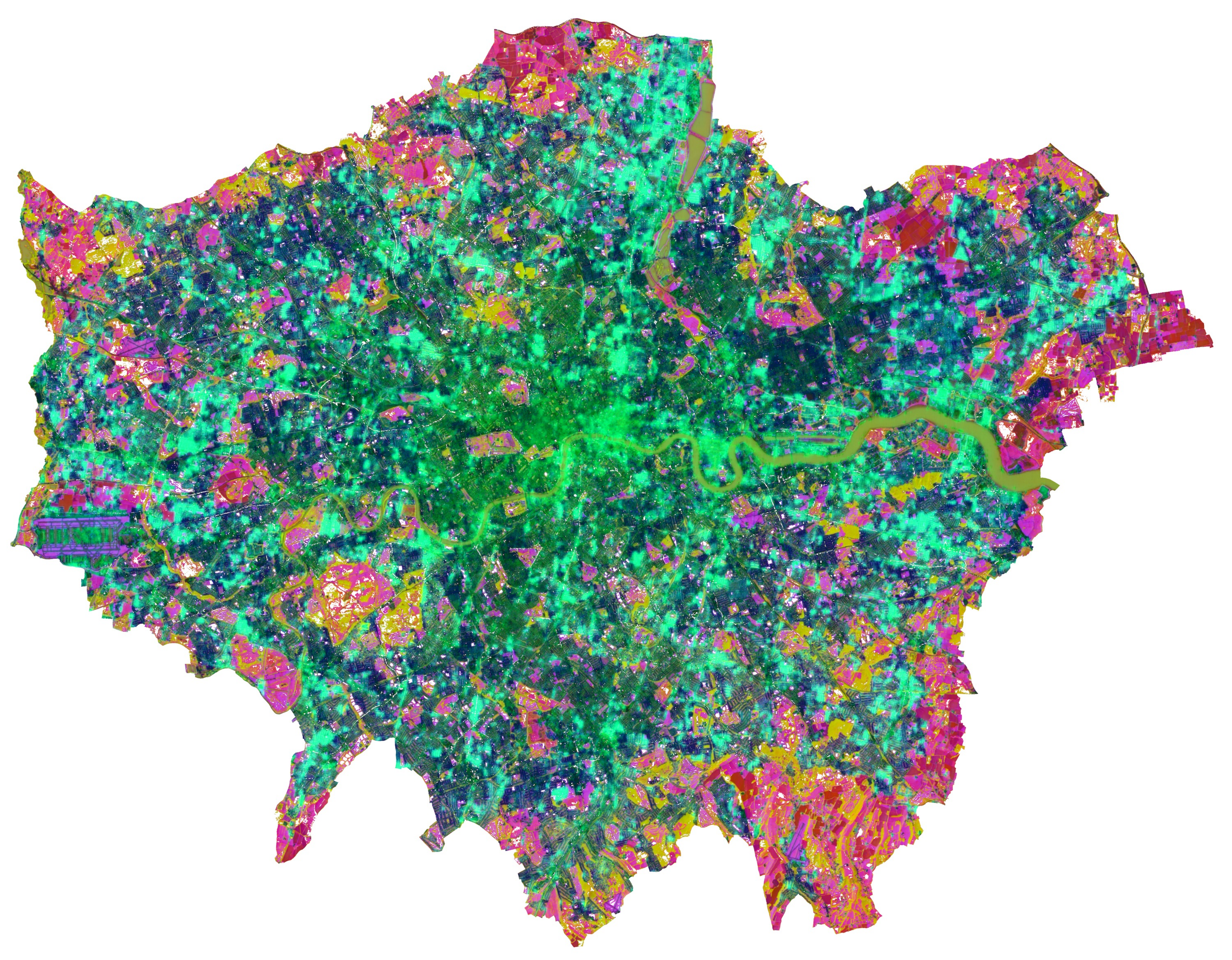}
    \caption{London -- AlphaEarth}
    \label{fig:ld_ae}
\end{subfigure}
\hfill
\begin{subfigure}[t]{0.23\linewidth}
    \centering
    \includegraphics[width=\linewidth]{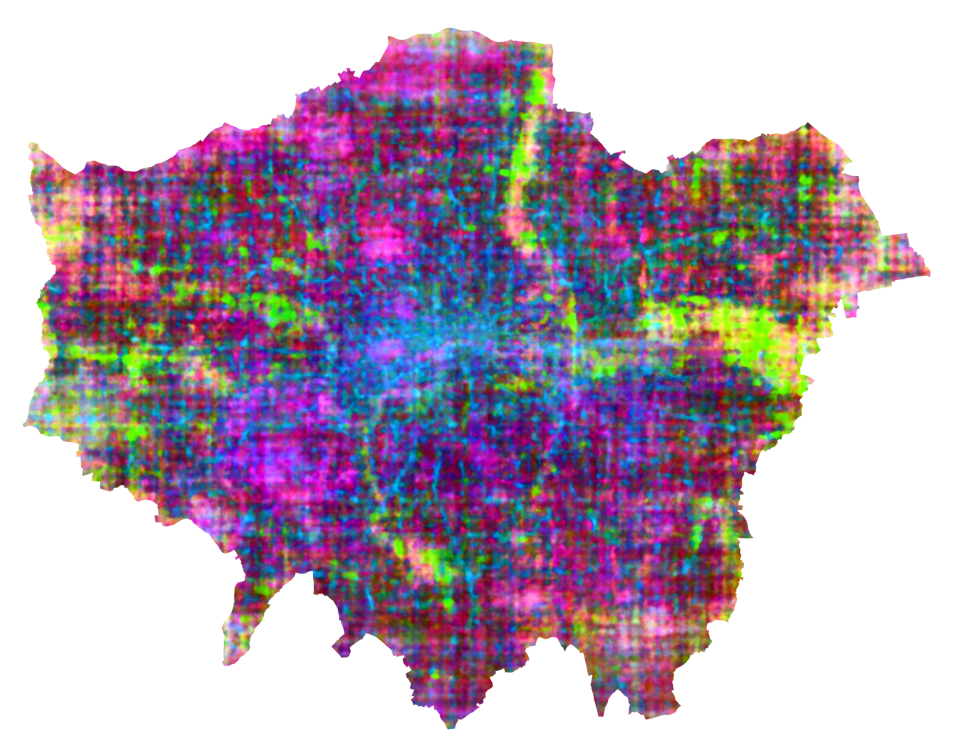}
    \caption{London -- CaLLiPer}
    \label{fig:ld_calliper}
\end{subfigure}
\hfill
\begin{subfigure}[t]{0.23\linewidth}
    \centering
    \includegraphics[width=\linewidth]{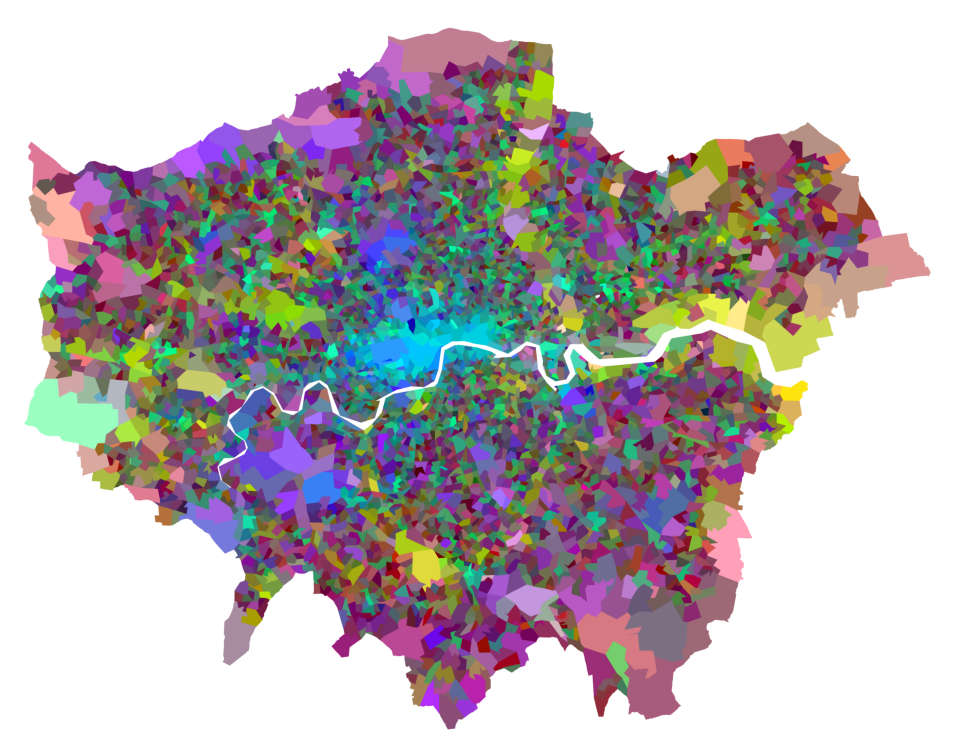}
    \caption{London -- LDA}
    \label{fig:ld_lda}
\end{subfigure}

\vspace{0.4cm}

\begin{subfigure}[t]{0.23\linewidth}
    \centering
    \includegraphics[width=\linewidth]{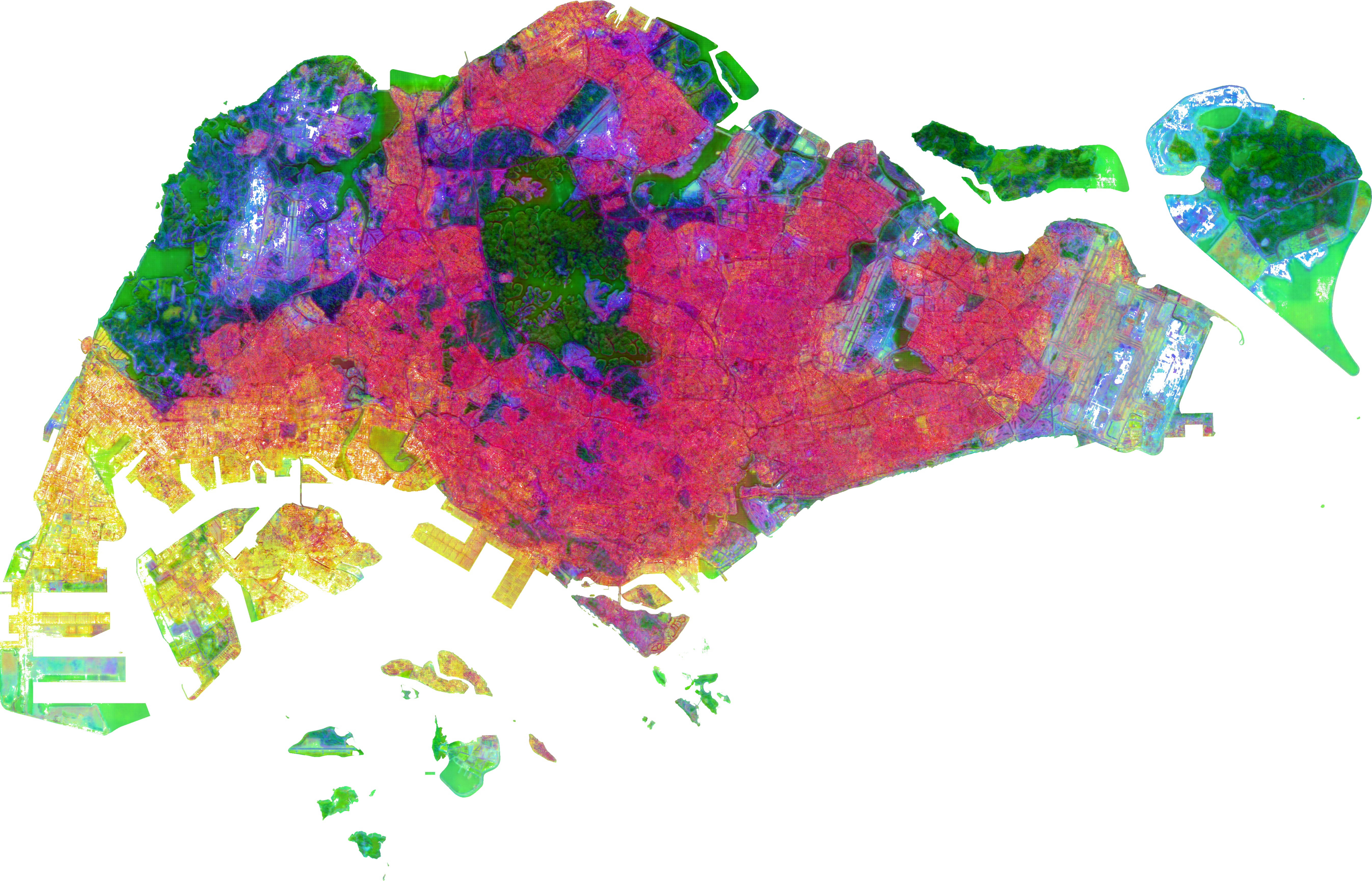}
    \caption{Singapore -- AETHER}
    \label{fig:sg_aether}
\end{subfigure}
\hfill
\begin{subfigure}[t]{0.23\linewidth}
    \centering
    \includegraphics[width=\linewidth]{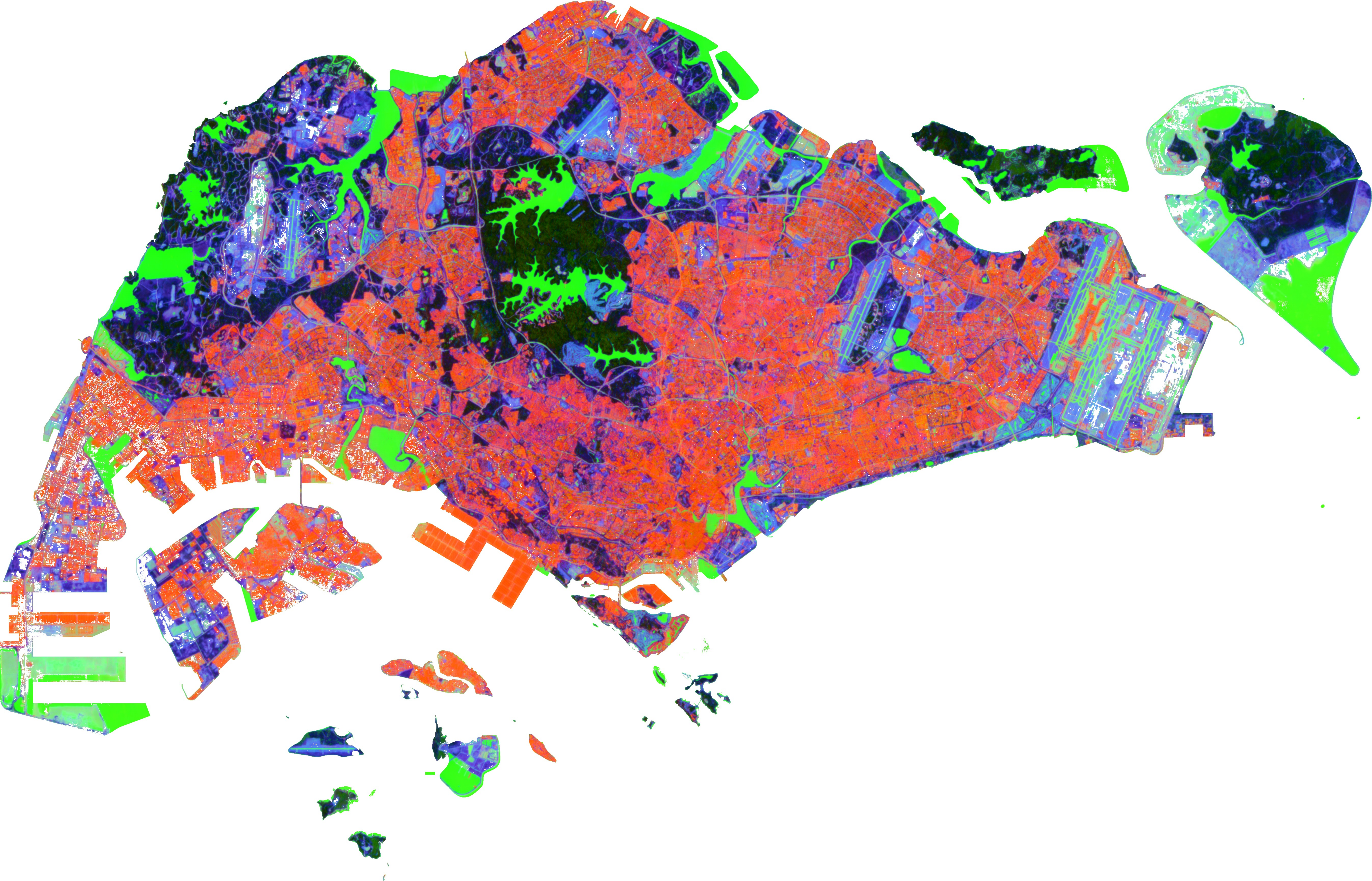}
    \caption{Singapore -- AlphaEarth}
    \label{fig:sg_ae}
\end{subfigure}
\hfill
\begin{subfigure}[t]{0.23\linewidth}
    \centering
    \includegraphics[width=\linewidth]{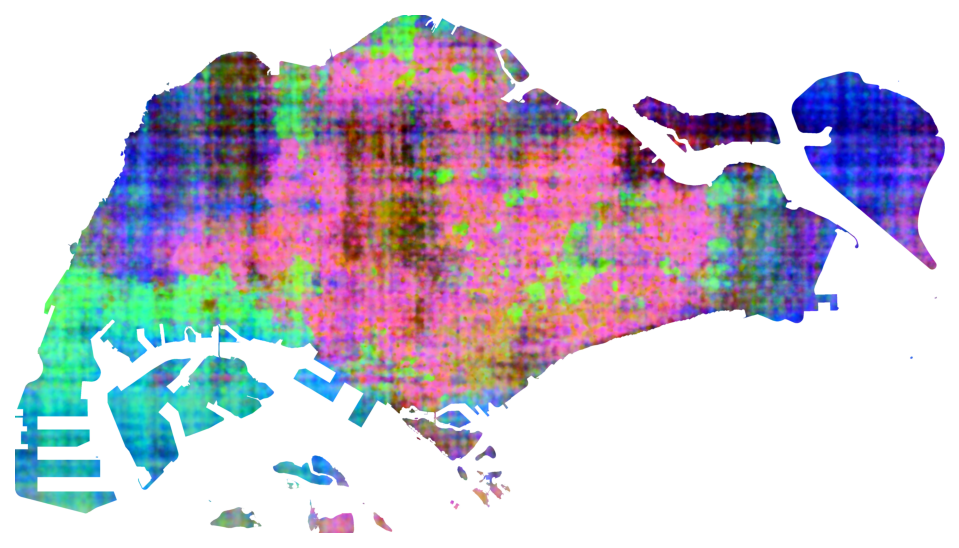}
    \caption{Singapore -- CaLLiPer}
    \label{fig:sg_calliper}
\end{subfigure}
\hfill
\begin{subfigure}[t]{0.23\linewidth}
    \centering
    \includegraphics[width=\linewidth]{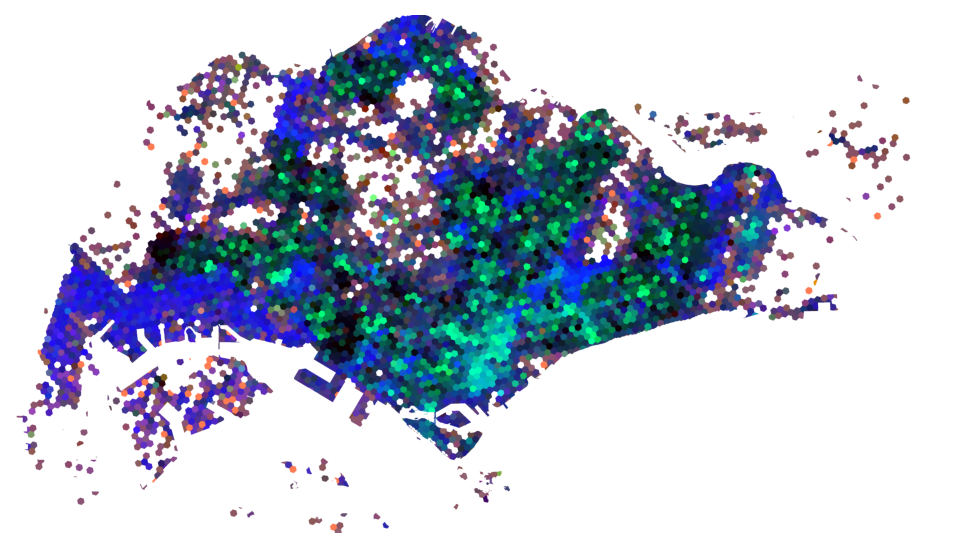}
    \caption{Singapore -- LDA}
    \label{fig:sg_lda}
\end{subfigure}

\caption{
PCA visualization of the first three principal components (RGB composite) of spatial embeddings across four models in London (top row) and Singapore (bottom row).
}
\label{fig:embedding_comparison}
\end{figure*}

Beyond downstream accuracy, we further investigate how POI-guided alignment reshapes the spatial embedding space. 
Figure~\ref{fig:embedding_comparison} visualizes the first three principal components of the city-wide embedding fields as RGB composites. 
We visualize four models to represent distinct paradigms of spatial representation learning:
AETHER (multimodal EO–POI alignment),
AlphaEarth (EO-based foundation model),
CaLLiPer (POI-driven semantic alignment), 
and LDA (traditional topic-based semantic modeling).
Across both cities, AETHER produces a more structured and internally differentiated spatial embedding structure. 

In London, the AETHER map (Fig.~\ref{fig:ld_aether}) reveals a clear urban structure.
Fine-grained linear patterns corresponding to the road network are more prominently preserved, forming coherent structural traces across the city.
At the same time, open and less-developed areas retain subtle morphological features.
This indicates that the aligned embedding emphasizes urban functions while preserving spatial structure.
By contrast, the AlphaEarth visualization (Fig.~\ref{fig:ld_ae}) exhibits a more block-like distribution composed of relatively homogeneous color regions.
The embedding field appears partitioned into large chromatic segments, suggesting that variance is dominated by broad spectral differences.
CaLLiPer (Fig.~\ref{fig:ld_calliper}) preserves the primary urban structural skeleton, including major road corridors and the river network.
However, its representation shows limited expressiveness in capturing detailed urban morphology, with reduced fidelity in fine-grained spatial shapes and transitions.
LDA (Fig.~\ref{fig:ld_lda}) retains coarse regional semantic segmentation, but lacks fine-grained differentiation, resulting in relatively homogeneous areas with limited internal structural variation.

In Singapore, AETHER (Fig.~\ref{fig:sg_aether}) shows richer color diversity and finer-grained variation within the urban core.
Instead of forming large uniform regions, the embedding space differentiates subtle intra-urban transitions, producing a more nuanced internal structure despite the city’s compact morphology.
In comparison, AlphaEarth (Fig.~\ref{fig:sg_ae}) is characterized by fewer dominant colors and more uniform blocks, with limited internal variation inside dense urban areas.
CaLLiPer in Singapore (Fig.~\ref{fig:sg_calliper}) presents a semantic distribution that reflects functional clustering but lacks detailed morphological delineation, resulting in spatial patterns that are semantically meaningful yet less shape-consistent.
LDA (Fig.~\ref{fig:sg_lda}) exhibits semantic differentiation over limited regions, but the number of distinguishable categories remains constrained, leading to coarse and spatially fragmented patterns.

Overall, the PCA composites indicate that AETHER not only improves predictive performance but also reorganizes the embedding field into a more expressive representation. Importantly, the results suggest that EO and POI signals are not merely superimposed in the aligned representation. 
Although POIs are introduced as discrete point-level supervision during training, the multi-scale module on physical-aware EO features enables semantic information to propagate beyond individual point locations, extending toward the spatial boundaries of corresponding urban objects.

\subsection{Spatial Mechanism Analysis}

\begin{figure*}[t]
    \centering
        \begin{subfigure}[t]{0.32\linewidth}
        \centering
        \includegraphics[width=\linewidth]
        {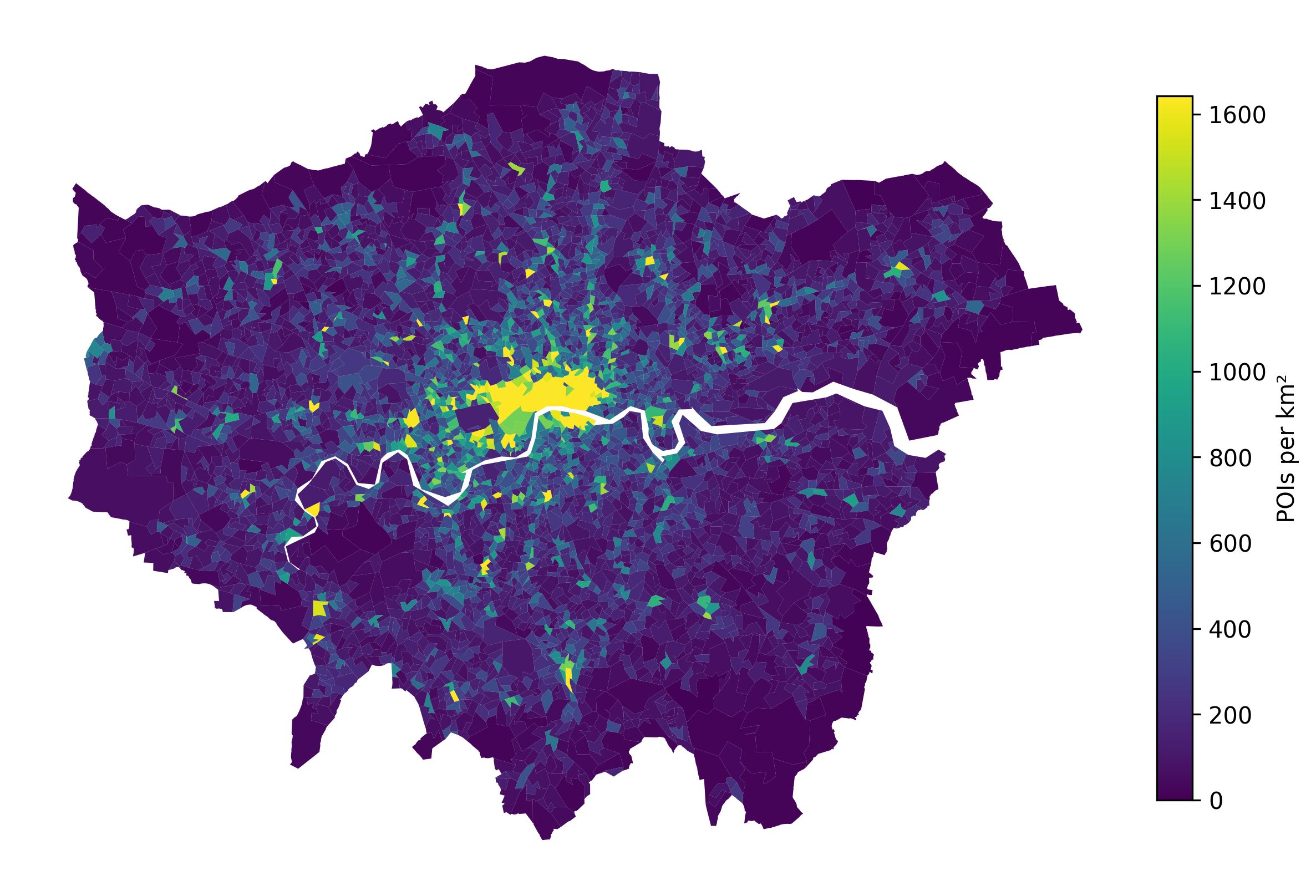}
        \caption{POI density (POIs per km$^{2}$)}
        \label{fig:poi_density}
        \hfill
    \end{subfigure}
    \hfill
    \begin{subfigure}[t]{0.32\linewidth}
        \centering
        \includegraphics[width=\linewidth]{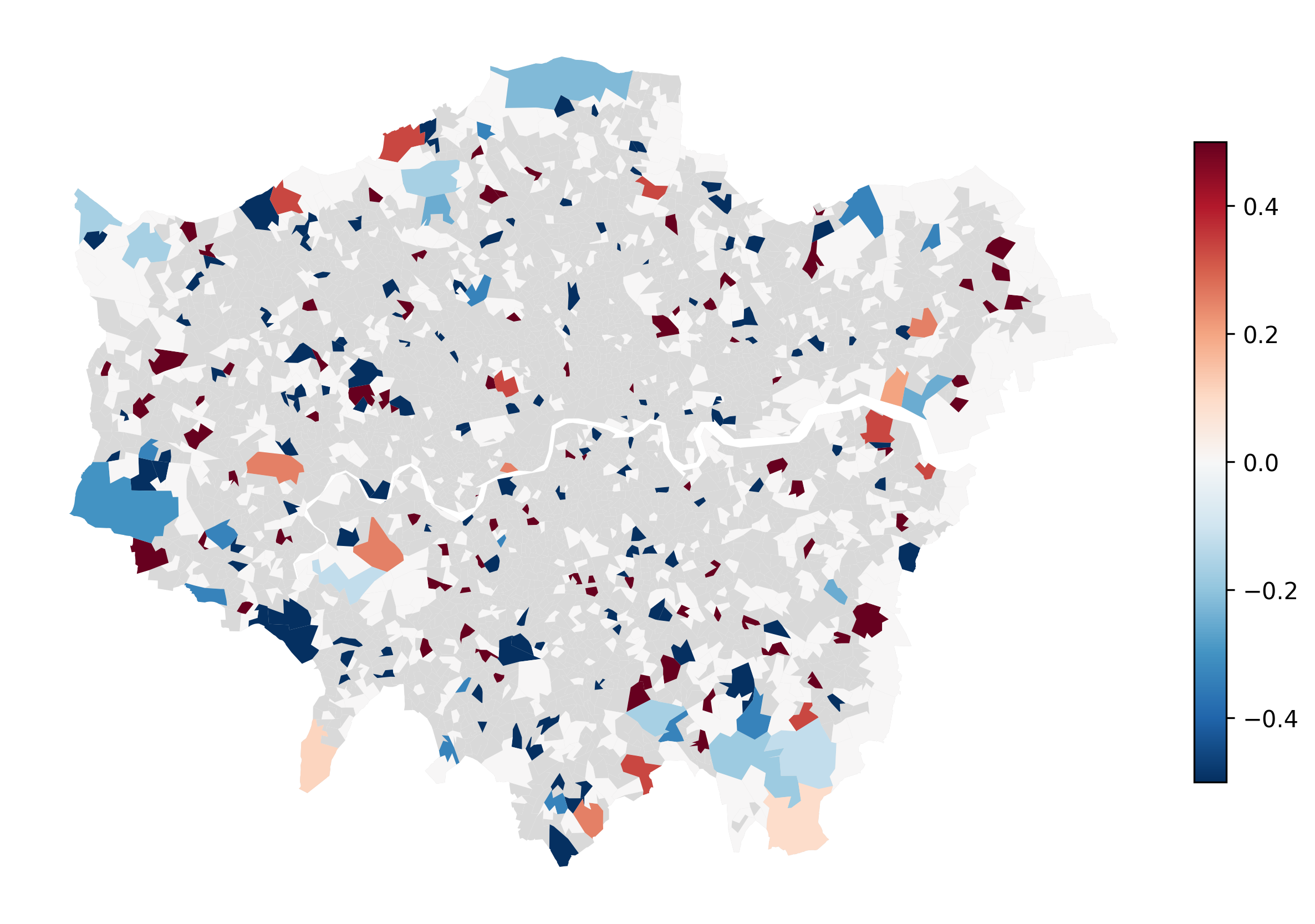}
        \caption{$\Delta$ LUC error rate (AETHER $-$ AlphaEarth)}
        \label{fig:luc_delta}
    \end{subfigure}
    \hfill
    \begin{subfigure}[t]{0.32\linewidth}
        \centering
        \includegraphics[width=\linewidth]
        {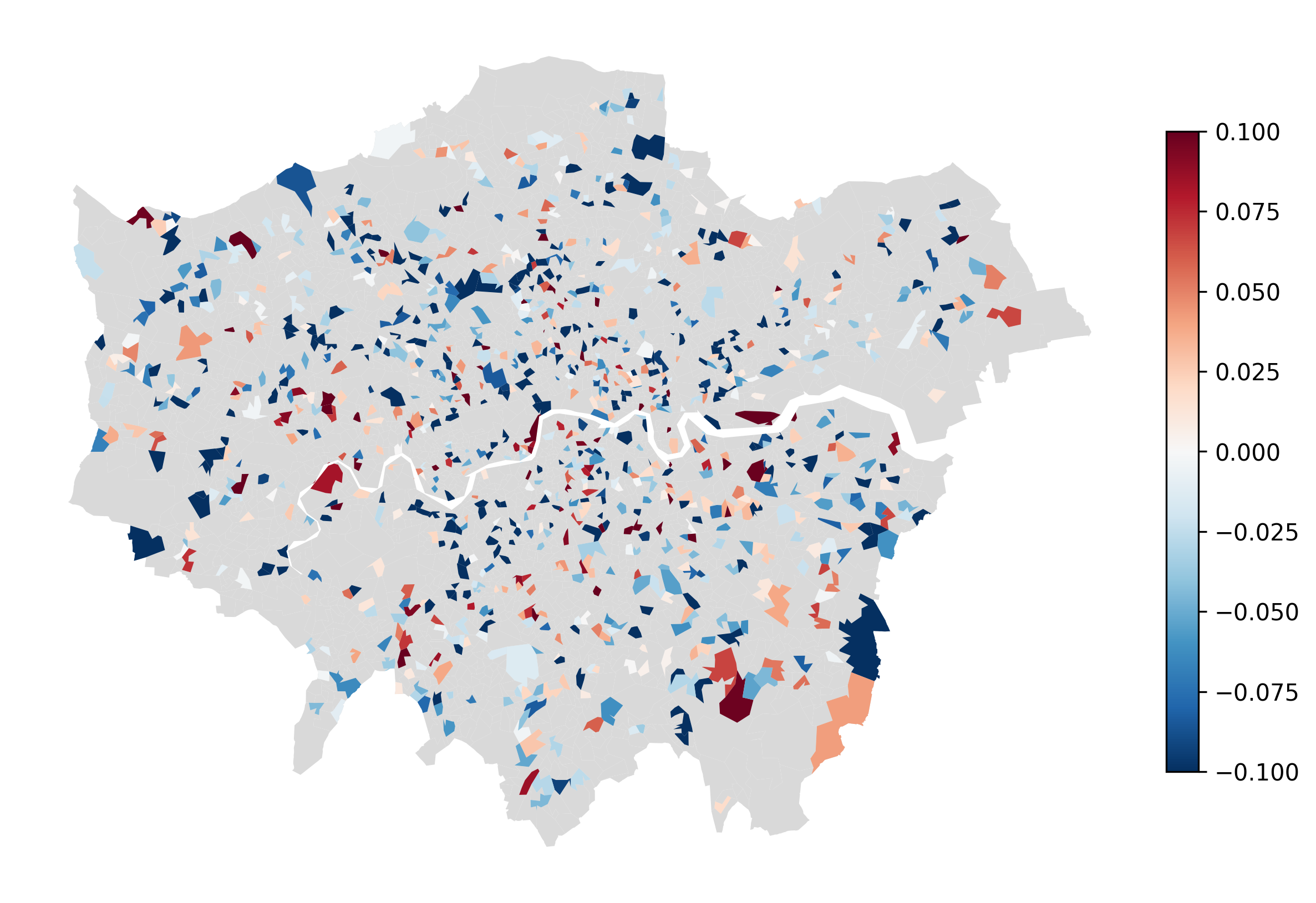}
        \caption{$\Delta$ SDM $L_{1}$ (AETHER $-$ AlphaEarth)}
        \label{fig:sdm_delta}
    \end{subfigure}
    \caption{
    Spatial relationship between POI density and model differences between AETHER and AlphaEarth in London.
    (a) POI density measured as the number of POIs per square kilometer.
    (b) Spatial distribution of land-use classification error difference.
    (c) Spatial distribution of socioeconomic distribution prediction difference ($L_{1}$ distance).
    Negative $\Delta$ values indicate lower error for AETHER compared with AlphaEarth.
    }
    \label{fig:spatial_mechanism}
\end{figure*}

To examine where improvements occur, Figure~\ref{fig:spatial_mechanism} maps the spatial difference between AlphaEarth and AETHER prediction errors.
Negative $\Delta$ values indicate reduced error under AETHER.
POI density (Panel a) is strongly concentrated in central London, with secondary clusters along major urban corridors.
For SDM (Panel b), larger error reductions are visible in central areas where POI density is highest.
This indicates that dense semantic supervision strengthens improvements in socioeconomic prediction.
However, negative $\Delta$ values are also observed across outer boroughs.
Improvements therefore extend beyond the urban core.
For LUC (Panel c), deeper reductions also appear in central regions.
At the same time, lighter but widespread improvements are distributed across peripheral areas.
Overall, both tasks exhibit a pattern of central amplification and city-wide diffusion.
While improvements are stronger in high-density POI regions, they are not restricted to them.
This suggests that semantic alignment enhances representations beyond individual POI clusters.

\begin{figure*}[t]
    \centering
    \begin{subfigure}[t]{0.45\linewidth}
        \centering
        \includegraphics[width=\linewidth]{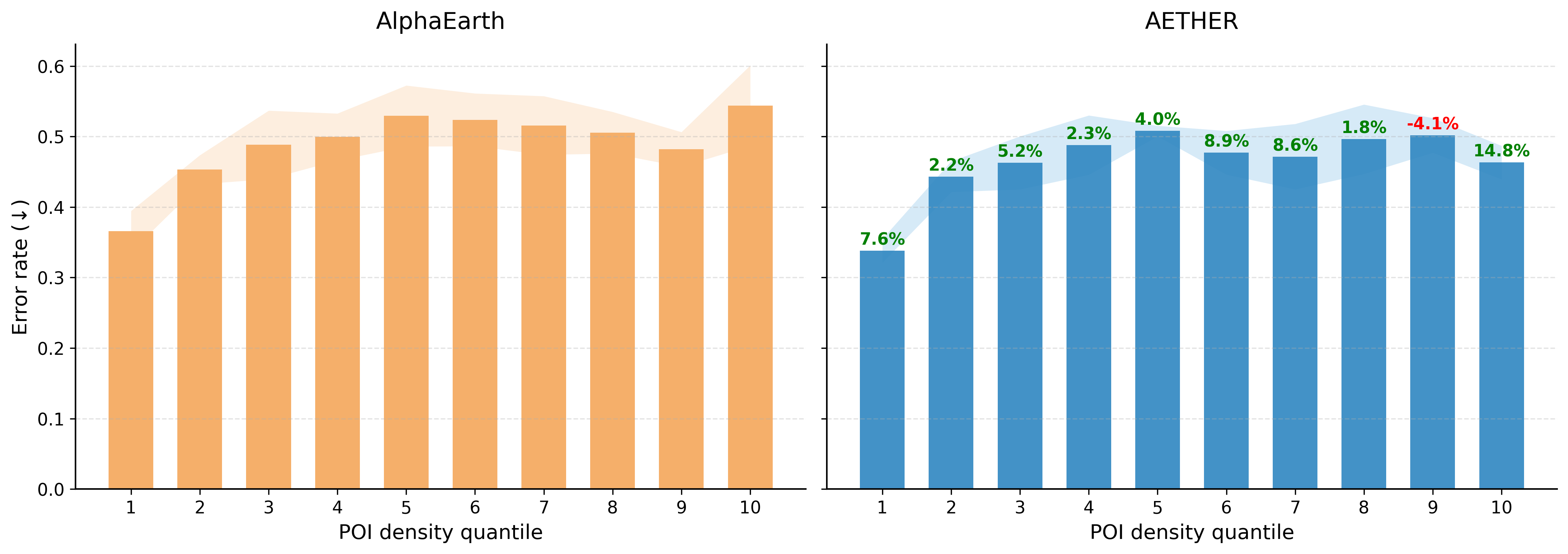}
        \caption{Comparison of LUC across POI density bins}
        \label{fig:luc_gain}
    \end{subfigure}
    \hfill
    \begin{subfigure}[t]{0.45\linewidth}
        \centering
        \includegraphics[width=\linewidth]{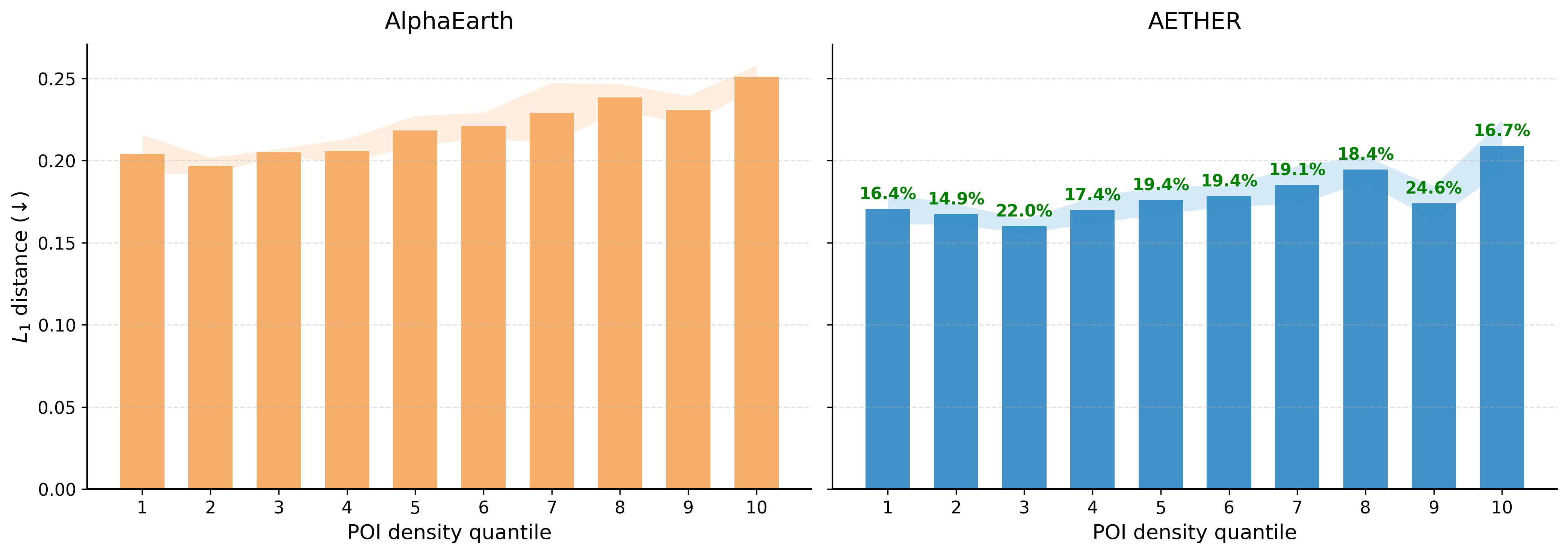}
        \caption{Comparison of SDM across POI density bins}
        \label{fig:sdm_gain}
    \end{subfigure}
    
    \caption{
    Performance comparison between AETHER and AlphaEarth across POI density quantile bins.
    }
    \label{fig:gain_by_density}
\end{figure*}

We further examine the effect of POI-guided alignment in Figure~\ref{fig:gain_by_density} using equal-frequency POI density bins. Across both tasks, AETHER improves over AlphaEarth in nearly all density bins. While the magnitude of improvement varies, several medium- and high-density bins exhibit particularly strong gains. This pattern indicates that the benefits of POI-guided alignment are broadly present across different POI density regimes rather than limited to a specific range of densities.
Variations in improvement magnitude likely reflect differences in both task difficulty and the amount of semantic information provided by POIs. Regions with higher POI density often correspond to functionally complex environments, where distinguishing between different functions in downstream tasks becomes more challenging. At the same time, richer POI signals provide additional semantic cues that help disambiguate functional patterns.
Overall, the improvements introduced by AETHER appear across regions with different POI densities, suggesting that the method is robust to variations in POI availability. At the same time, stronger gains in some POI-dense areas indicate that richer semantic supervision can further enhance functional inference.

\subsection{Natural Language–Conditioned Spatial Retrieval}
\label{sec:results:retrieval}

\begin{figure*}[t]
    \centering
    \includegraphics[width=\linewidth]{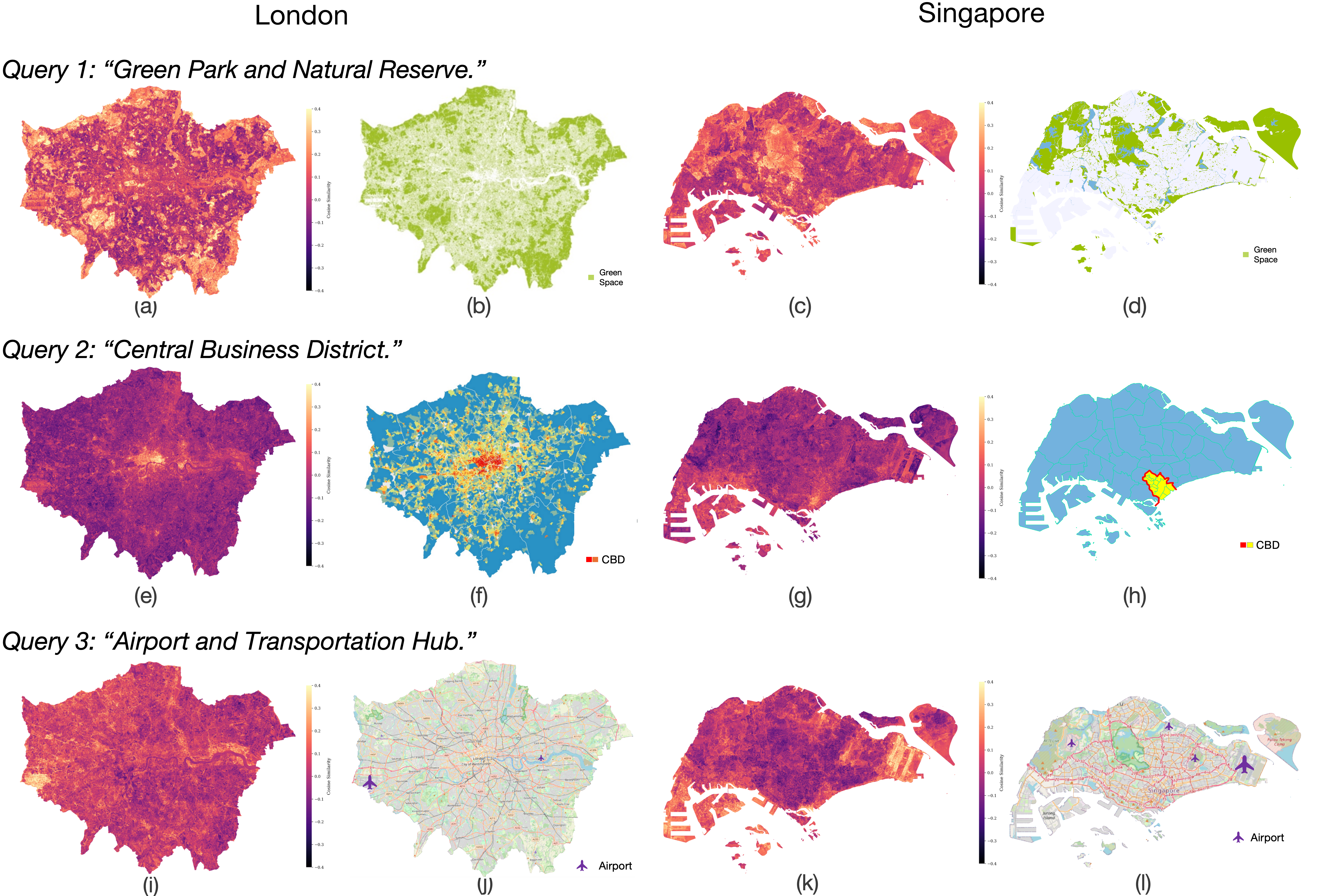}
    \caption{
    Spatial retrieval results using AETHER embeddings under three natural language prompts.
    For each city, cosine similarity maps (left panels) are compared against ground truth references (right panels) derived from authoritative planning documents and open geospatial datasets.
    Airport ground truth is represented as point locations, whereas CBD and green spaces are polygon-based.
    The color scale indicates cosine similarity values, with warmer colors representing stronger semantic correspondence. 
    }
    \label{fig:retrieval}
\end{figure*}

Beyond supervised prediction, we evaluate whether the aligned embedding space supports direct natural language querying of urban space. 
Given a free-form textual prompt, we compute cosine similarity between the query embedding and all spatial embeddings, producing a full-coverage semantic response map (Figure~\ref{fig:retrieval}). Similarity maps are computed at the original AE spatial resolution (10 m grid).

\paragraph{Natural language prompt localization}
Figure~\ref{fig:retrieval} presents spatial retrieval results under three natural language queries in London and Singapore. 
Ground-truth spatial references are compiled from official planning datasets and open geospatial sources.
For London, green-space layers are obtained from the London Datastore green cover dataset.
The spatial extent of the London CBD is interpreted based on employment-density patterns reported in GLA economic working papers \citep{girardi2017london_economy}.
For Singapore, CBD boundaries follow the official Downtown Core planning area defined by the Urban Redevelopment Authority (URA), and green spaces are inferred from the URA Master Plan land-use dataset.
Airport locations and major transportation infrastructure are derived from OpenStreetMap.

For Query 1 (``Green Park and Natural Reserve''), green spaces in both cities are accurately highlighted.
The response maps not only localize major parks and natural reserves, but also recover their spatial extent and boundary structure.
In London, parks such as Hyde Park and Richmond Park are clearly delineated.
In Singapore, nature reserves and coastal green areas are distinctly emphasized.
Although POI supervision during training is point-based, the aligned embedding successfully captures polygon-level spatial structure already encoded in the AE features.
This indicates that semantic alignment enhances region-level functional representation rather than merely activating individual POI points.

For Query 2 (``Central Business District''), high similarity scores concentrate in the known urban cores.
In London, the central financial area around the City of London and Canary Wharf is strongly highlighted.
In Singapore, the CBD region in the southern central area is sharply localized.
The response maps exhibit compact and coherent spatial clusters, demonstrating that the embedding space captures concentrated economic functions.

For Query 3 (``Airport and Transportation Hub''), major transportation infrastructures are distinctly identified.
In London, Heathrow Airport is prominently highlighted, along with major rail hubs such as Stratford and transportation corridors along the Thames.
In Singapore, Changi Airport, several military air bases, and the ports are clearly emphasized.
The retrieval maps show focused responses around large-scale transportation facilities rather than diffuse activation.

Overall, the qualitative results demonstrate that AETHER supports open-vocabulary spatial localization well.
The model captures both semantic meanings and the spatial contours of functional regions.
These findings indicate that POI-guided alignment grounds EO embeddings in language while preserving spatial continuity and structural detail.

\paragraph{Quantitative retrieval}

Table~\ref{tab:poi_recall_final} reports quantitative retrieval performance across five representative POI categories using the Recall@Top-$k\%$ metric.
The evaluation measures the proportion of ground-truth POI locations that fall within the top-$k\%$ highest-scoring spatial pixels ranked by cosine similarity.

Across all categories and thresholds, AETHER consistently achieves the best retrieval performance.
At the strict Top 1\% threshold, AETHER substantially outperforms the baselines, reaching 54.56\% recall for Arts \& Culture and 14.91\% for Medical locations, compared with 16.39\% and 4.16\% achieved by CaLLiPer.
The performance gap remains large across other categories, indicating that the aligned embedding space provides strong semantic localization capability even under highly selective retrieval conditions.

As the retrieval threshold increases, recall improves for all methods.
However, AETHER maintains a clear advantage at both Top 5\% and Top 10\%.
These results demonstrate that the AETHER embedding space enables accurate semantic localization across diverse urban functions.
By integrating EO-derived morphological signals with POI-based semantic supervision, the aligned representation captures both physical context and functional meaning.

\paragraph{Implications}
Together, the qualitative and quantitative results demonstrate that multimodal alignment transforms AE embeddings into language-accessible spatial representations. 
As a result, urban regions can be queried using natural language descriptions. 
This capability extends beyond traditional supervised mapping and represents a step toward more interpretable geospatial foundation models.

\begin{table*}[t]
\centering
\setlength{\tabcolsep}{6pt}
\renewcommand{\arraystretch}{1.15}
\caption{
Recall@Top-$k\%$ (\%) comparison across five POI categories.
Abbreviations: Art = Art and Antiques; 
Med. = Medical Equipment, Supplies and Pharmaceuticals; 
Imp. = Import and Export Services; 
Sport = Sports Clubs and Associations; 
Rel. = Religious Organisations.
AlphaEarth is trained without text alignment; retrieval under the text-based protocol is therefore not applicable (---).
}
\label{tab:poi_recall_final}
\begin{tabular}{llccccc}
\toprule
\textbf{Metric} & \textbf{Method}
& \textbf{Art} 
& \textbf{Med.} 
& \textbf{Imp.} 
& \textbf{Sport} 
& \textbf{Rel.} \\
\midrule

\multirow{4}{*}{Top 1\%}
& Random       & 0.41  & 0.98  & 0.83  & 1.90  & 0.44  \\
& CaLLiPer     & 16.39 & 4.16  & 2.70  & 2.95  & 1.97  \\
& AlphaEarth   & ---   & ---   & ---   & ---   & ---   \\
& AETHER       & \textbf{54.56} & \textbf{14.91} & \textbf{12.45} & \textbf{6.33}  & \textbf{6.99} \\
\midrule

\multirow{4}{*}{Top 5\%}
& Random       & 4.15  & 4.16  & 6.22  & 6.33  & 3.71  \\
& CaLLiPer     & 28.22 & 12.22 & 12.03 & 17.09 & 15.72 \\
& AlphaEarth   & ---   & ---   & ---   & ---   & ---   \\
& AETHER       & \textbf{72.82} & \textbf{48.41} & \textbf{37.14} & \textbf{22.15} & \textbf{18.56} \\
\midrule

\multirow{4}{*}{Top 10\%}
& Random       & 8.71  & 10.51 & 9.75  & 12.24 & 8.95  \\
& CaLLiPer     & 34.65 & 17.60 & 20.95 & 27.85 & 26.64 \\
& AlphaEarth   & ---   & ---   & ---   & ---   & ---   \\
& AETHER       & \textbf{79.46} & \textbf{69.68} & \textbf{49.59} & \textbf{32.07} & \textbf{29.69} \\
\bottomrule
\end{tabular}
\end{table*}

\subsection{Ablation and Sensitivity Analysis}
\label{sec:res:sensitivity}

\paragraph{Ablation Study}

Table~\ref{tab:london_ablation} reports the ablation results under four controlled variants:
(i) direct concatenation of AE embeddings and mean-aggregated POI text embeddings at each spatial unit, 
(ii) removing the intra-modal AE–AE consistency loss ($\lambda=0$),
(iii) excluding POI name information from textual descriptions, and
(iv) replacing the original text encoder with the SentenceTransformer model \texttt{all-MiniLM-L6-v2}.

Naive concatenation leads to clear performance degradation, especially for SDM,
indicating that simple feature fusion is insufficient and that contrastive alignment is essential for semantic integration.
Removing the AE–AE consistency term results in only minor performance drops,
suggesting that it mainly serves as a regularizer as designed.
Excluding POI name information reduces both LUC and SDM performance,
highlighting the contribution of fine-grained textual semantics.
Replacing the text encoder yields comparable but slightly weaker results, confirming the robustness of the proposed design.

\begin{table*}[t]
\centering
\small
\setlength{\tabcolsep}{6pt}
\caption{
Ablation study in London.
All results are reported as mean $\pm$ std over five random seeds.
}
\label{tab:london_ablation}
\begin{tabular}{l|ccc|ccc}
\toprule
\multirow{2}{*}{Variant} &
\multicolumn{3}{c|}{LUC (↑)} &
\multicolumn{3}{c}{SDM (↓)} \\
\cmidrule(lr){2-4}\cmidrule(lr){5-7}
& F1 & Precision & Recall &
KL & L1 & Chebyshev \\
\midrule

AETHER (ours) 
& \textbf{59.4 $\pm$ 0.9} 
& \textbf{58.3 $\pm$ 1.0} 
& \textbf{61.8 $\pm$ 0.9} 
& \textbf{30.1 $\pm$ 1.2} 
& \textbf{17.8 $\pm$ 0.3} 
& \textbf{53.8 $\pm$ 1.1} \\

Concat (AE + POI text embedding)
& 56.9 $\pm$ 1.2
& 56.7 $\pm$ 1.2
& 58.6 $\pm$ 1.3
& 41.9 $\pm$ 1.1
& 21.6 $\pm$ 0.3
& 64.5 $\pm$ 0.7 \\

w/o AE–AE consistency
& 59.0 $\pm$ 0.8
& 57.9 $\pm$ 0.8
& 61.5 $\pm$ 1.0
& 31.1 $\pm$ 1.0
& 18.2 $\pm$ 0.3
& 54.5 $\pm$ 1.2 \\

w/o POI name
& 58.6 $\pm$ 0.2
& 57.5 $\pm$ 0.5
& 61.2 $\pm$ 0.4
& 32.8 $\pm$ 0.8
& 18.7 $\pm$ 0.3
& 56.2 $\pm$ 0.8 \\

SentenceTransformer text encoder
& 59.1 $\pm$ 0.5
& 57.8 $\pm$ 0.7
& 61.5 $\pm$ 0.4
& 31.9 $\pm$ 1.3
& 18.3 $\pm$ 0.4
& 55.2 $\pm$ 1.2 \\

\bottomrule
\end{tabular}
\end{table*}

\paragraph{Effect of loss balance ($\lambda$)}

\begin{figure}[t]
    \centering
    
    \begin{subfigure}[t]{\linewidth}
        \centering
        \includegraphics[width=\linewidth]{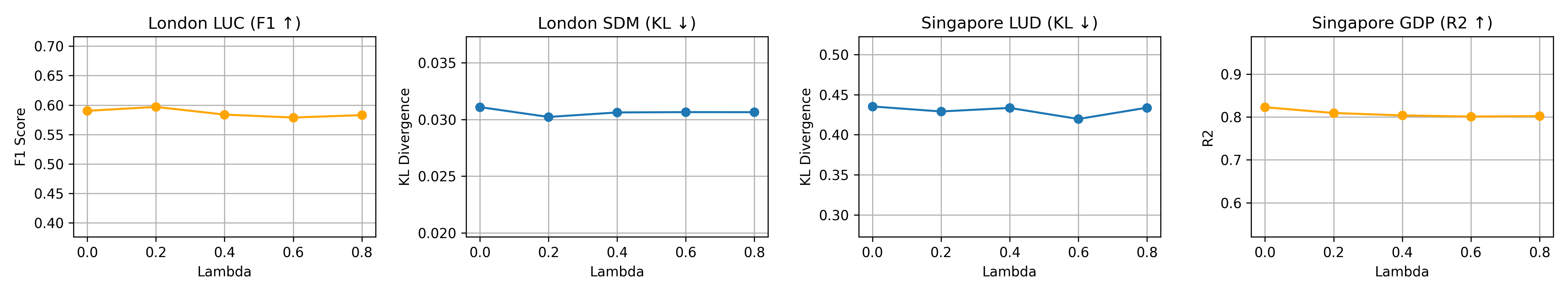}
        \caption{Sensitivity to the loss balance coefficient $\lambda$.}
        \label{fig:sensitivity_lambda}
    \end{subfigure}
    
    \vspace{0.4cm}
    
    \begin{subfigure}[t]{\linewidth}
        \centering
        \includegraphics[width=\linewidth]{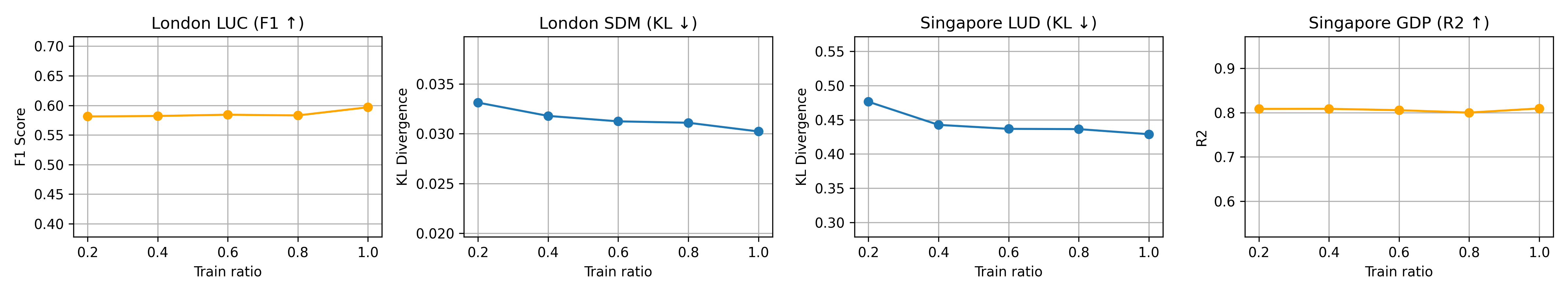}
        \caption{Model performance under varying training data volumes.}
        \label{fig:sensitivity_train}
    \end{subfigure}
    
    \caption{Sensitivity analysis of AETHER.}
    \label{fig:sensitivity_combined}
\end{figure}

Figure~\ref{fig:sensitivity_lambda} shows model sensitivity to the loss balance coefficient $\lambda$.
In London, the best performance is achieved around $\lambda = 0.2$ across the evaluated tasks.
When $\lambda$ increases, performance on both tasks gradually declines, indicating that large-scale AE self-alignment serves primarily as an auxiliary signal.  
In Singapore, performance remains relatively stable across different $\lambda$ values.
This lower sensitivity may relate to the city’s more compact and homogeneous urban structure, where semantic signals from POIs are spatially denser and more consistent.
Overall, the results indicate that AETHER performs robustly.

\paragraph{Effect of training data volume}
Figure~\ref{fig:sensitivity_train} evaluates model performance under varying proportions of POI–AE training pairs.
For LUC and GDP prediction, performance remains relatively stable as the training ratio changes.
This suggests that these tasks may rely strongly on EO-derived morphological information already captured in the AE embeddings, or that the related alignment can be learned easily.
In contrast, SDM and LUD show consistent improvements as more training data become available.
These tasks depend more heavily on semantic signals related to complex human activities and socioeconomic functions.
Increasing the amount of POI supervision therefore helps the model better align spatial embeddings with functional urban patterns.

\subsection{Efficiency and Scalability}
\label{sec:exp:efficiency}

Table~\ref{tab:efficiency_comparison} compares model complexity and per-epoch alignment training time under the same hardware configuration (NVIDIA RTX PRO 6000, 96GB).
AETHER contains only 0.17M trainable parameters and requires approximately 4.7 seconds per epoch, demonstrating competitive efficiency among lightweight spatial representation models.
Despite incorporating multimodal contrastive alignment, its training cost remains comparable to or lower than existing baselines.
During inference, embeddings are generated by applying a lightweight projection head to precomputed AE features, resulting in negligible additional computational overhead.

\begin{table}[t]
\centering
\caption{
Comparison of model complexity and per-epoch training time under the same hardware configuration (NVIDIA RTX PRO 6000, 96GB).
}
\label{tab:efficiency_comparison}
\setlength{\tabcolsep}{6pt}
\renewcommand{\arraystretch}{1.15}
\begin{tabular}{lcc}
\toprule
Model
& Trainable Params (M) 
& Time / Epoch (s) \\
\midrule
Place2Vec & 0.06  & 4.87 \\
SPPE      & 0.12  & 40.04 \\
Space2Vec & 0.68  & 10.98 \\
CaLLiPer  & 0.69  & 7.10 \\
\midrule
AETHER (ours) & 0.17 & 4.72 \\
\bottomrule
\end{tabular}
\end{table}

\section{Discussion}

This section discusses the broader implications of the results for urban representation learning with geospatial foundation models. 
We examine the complementary roles of EO and POI signals, how semantic alignment reshapes the spatial embedding space, and the broader implications of language-aligned spatial representations, followed by a discussion of current limitations and future directions.

\paragraph{Complementarity between EO and POI semantics}

Our experiments consistently show that EO-derived embeddings and POI-based representations capture different aspects of urban space. AlphaEarth embeddings already provide strong performance for physically grounded tasks such as land-use classification in London, where spectral and structural patterns are highly informative. However, for socially grounded tasks such as socioeconomic distribution mapping and GDP prediction, EO signals alone are insufficient. POI data offer complementary information by encoding functional semantics associated with human activities and institutions. By aligning AE embeddings with POI textual representations, AETHER integrates these two sources of information within a unified latent space. The resulting representation therefore combines the spatial completeness of EO observations with the functional richness of urban semantics. This integration explains why AETHER improves performance on both physically grounded and functionally grounded tasks while maintaining spatial coherence.

\paragraph{Semantic restructuring of spatial embedding space}

Beyond improvements in predictive accuracy, our analysis suggests that POI-guided contrastive alignment reshapes the structure of the embedding field. 
The PCA visualizations indicate that AETHER produces embeddings with richer internal differentiation and clearer spatial organization compared with the original AlphaEarth features. This suggests that semantic alignment reorganizes the latent feature space so that locations with similar urban functions become closer in the embedding manifold. Importantly, improvements are not limited to locations with dense POI supervision. Both the spatial error maps and the density-stratified analysis show that performance gains extend to suburban and peripheral regions where POIs are sparse. This indicates that semantic information learned from POIs propagates through the continuous AE embedding field, enabling the model to generalize functional signals beyond discrete training anchors. Such propagation is particularly valuable for urban areas where semantic data coverage is uneven.

\paragraph{Toward language-accessible geospatial foundation models}

Another important implication of this work is the emergence of language-accessible spatial representations grounded in multimodal integration. By aligning EO embeddings with textual POI semantics, AETHER enables natural language–conditioned spatial retrieval, supporting open-vocabulary queries that localize functional regions such as central business districts, transportation hubs, and green spaces. Beyond retrieval performance, the framework illustrates a structured mechanism for integrating heterogeneous spatial signals. EO embeddings provide morphology-aware and spatially continuous features, while POI text introduces discrete yet semantically rich supervision. Rather than simply concatenating modalities, AETHER employs a multi-scale alignment design in which point-level semantic signals are modulated by underlying morphological structure and propagated toward spatially coherent object boundaries. The resulting embedding space jointly organizes human-centered semantics and physically grounded EO features, bridging latent representations and interpretable queries and suggesting a pathway toward interactive geospatial foundation models driven by natural language.

\paragraph{Limitations and future directions}

Despite these promising results, several limitations remain. 
First, POI data exhibit strong spatial heterogeneity and uneven coverage across regions. 
The semantic categories, naming conventions, and density of POIs can vary substantially between cities and countries, which may affect the stability of semantic alignment. 
While this study focuses on two metropolitan regions, extending the framework to large-scale cross-city or global settings remains an important direction for future research.
Second, the current framework relies primarily on EO embeddings and POI semantics. 
Other urban data modalities, such as street-level imagery and human mobility traces, also contain valuable signals about urban functions and human activities. 
Incorporating these complementary modalities may further enrich spatial representations and improve semantic grounding.
Third, the natural language–conditioned retrieval experiments presented here provide an initial demonstration of language-aligned spatial embeddings. 
Future work may explore more advanced retrieval benchmarks, improved semantic alignment strategies, and broader applications of language-accessible geospatial representations in urban analytics and interactive GIS systems.

\paragraph{Potential}
Beyond the evaluated tasks, the proposed framework opens up several potential applications in urban analytics and geospatial intelligence. Because AETHER produces semantically enriched spatial embeddings with full spatial coverage, the learned representation can support a wide range of downstream analyses, including urban functional zone mapping, infrastructure accessibility assessment, and urban planning support. In addition, the ability to query spatial embeddings using natural language provides an intuitive interface, enabling interactive exploration of urban environments through semantic descriptions. Such capabilities may facilitate new forms of human–AI interaction in geographic information systems, where complex spatial patterns can be discovered and interpreted through flexible language queries rather than predefined analytical pipelines.

\section{Conclusion}

This study introduces AETHER, a POI-guided multimodal contrastive alignment framework that adapts AlphaEarth embeddings for human-centered urban understanding. 
By aligning EO-derived spatial embeddings with textual representations of Points of Interest, AETHER integrates physically grounded EO features with functional urban semantics within a shared embedding space.
Experiments across two metropolitan regions and four representative urban tasks show that the aligned representations consistently outperform existing baselines. 
Beyond predictive improvements, the learned embedding space also supports natural language–conditioned spatial retrieval, enabling open-vocabulary localization of urban functions.
These results highlight the value of combining complementary spatial signals: EO data provide spatial completeness and morphology-aware context, while POI semantics introduce functional meaning associated with human activities. 
More broadly, this work demonstrates how physically grounded EO foundation models can be extended toward semantically enriched and language-accessible spatial representations, contributing to the development of more general and interpretable geospatial foundation models.

\section*{Declaration of Generative AI and AI-assisted Technologies in the Writing Process}

During the preparation of this manuscript, the authors used generative AI tools for language editing and text refinement. The authors reviewed and edited the content as needed and take full responsibility for the content of the published article.
\bibliography{cas-refs}

\end{document}